\documentclass{article}

 \usepackage[preprint]{neurips_2025}

\usepackage[utf8]{inputenc} 
\usepackage[T1]{fontenc}    
\usepackage{hyperref}       
\usepackage{url}            
\usepackage{booktabs}       
\usepackage{amsfonts}       
\usepackage{nicefrac}       
\usepackage{microtype}      
\usepackage{arydshln}

\usepackage{microtype}
\usepackage{graphicx}

\usepackage{amsmath}
\usepackage{amssymb}
\usepackage{mathtools}
\usepackage{amsthm}

\usepackage[capitalize,noabbrev]{cleveref}

\theoremstyle{plain}
\newtheorem{theorem}{Theorem}[section]

\theoremstyle{definition}

\theoremstyle{remark}

\usepackage{bm}
\usepackage{url}
\usepackage{pifont}
\usepackage{graphicx}
\usepackage{enumitem}
\usepackage{pgfplots}
\usepackage{xspace}
\usepackage{xcolor}
\usepackage{pgf-pie}
\usepackage{multirow}
\usepackage{multicol}
\usepackage{graphicx}
\usepackage{enumitem}
\usepackage{wrapfig}
\usepackage{lipsum}
\usepackage{subcaption}
\usepackage{nicematrix}

\pgfplotsset{compat=1.18}
\usepgfplotslibrary{polar}
\usetikzlibrary{intersections}
\usepgfplotslibrary{fillbetween}
\usetikzlibrary{backgrounds, calc}

\DeclareMathOperator*{\argmax}{arg\,max}

\definecolor{softred}{RGB}{250,100,100}
\definecolor{softgreen}{RGB}{56,118,29}
\definecolor{softblue}{RGB}{100,150,200}
\definecolor{softgray}{RGB}{150,150,150}
\newcommand{\textred}[1]{{\color{softred}#1}}
\newcommand{\textblue}[1]{{\color{softblue}#1}}

\usepackage{xspace}
\newcommand{\ourmetric}{\textsc{LCS}\xspace}

\title{Learning-to-Context Slope: Evaluating In-Context Learning Effectiveness Beyond Performance Illusions}

\author{
    \textbf{Dingzirui Wang$^{1}$ \quad Xuanliang Zhang$^{1}$ \quad Keyan Xu$^{1}$ \quad Qingfu Zhu$^{1}$} \\ \textbf{Wanxiang Che$^{1}$ \quad Yang Deng$^{2}$}\\
    $^{1}$Harbin Institute of Technology \quad $^{2}$Singapore Management University \\
    \texttt{\{dzrwang, xuanliangzhang, kyxu, qfzhu, car\}@ir.hit.edu.cn} \\
    \texttt{ydeng@smu.edu.sg}
}

\usepackage{etoolbox}

\begin{document}
    \maketitle
    \begin{abstract}
      In-context learning (ICL) has emerged as an effective approach to enhance the performance of large language models (LLMs). 
      However, its effectiveness varies significantly across models and tasks, posing challenges for practitioners to determine when ICL reliably improves performance. 
      Current evaluation approaches, reliant on performance change after applying ICL, suffer from low reliability, poor attribution, and impracticality in data-insufficient scenarios. 
      We propose the \textbf{Learning-to-Context Slope (\ourmetric)}, a novel metric that quantifies ICL effectiveness by modeling the slope between \textit{learning gain} (loss decrease from demonstrations) and \textit{contextual relevance} (demonstration-input relevance). 
      \ourmetric addresses key limitations of performance-based metrics: 
      (1) it captures continuous loss changes even when outputs are incorrect, improving reliability; 
      (2) its formulation attributes ICL failures to weak contextual alignment (inability to adapt inputs to demonstrations) or strong output calibration (self-verification of correctness); and 
      (3) it minimizes reliance on labeled data via synthetic evaluation. Extensive experiments demonstrate that \ourmetric strongly correlates with performance improvements in labeled settings and reliably reflects true effectiveness in biased or data-scarce scenarios. 
      Further analysis reveals actionable thresholds for \ourmetric and identifies model capabilities critical to ICL success\footnote{Our code and data will be released upon acceptance.}.
    \end{abstract}

    \section{Introduction}
        In-context learning (ICL) has emerged as a popular and effective paradigm for enhancing large language model (LLM) performance across diverse tasks, as it eliminates the need to retrain the LLMs \cite{brown-etal-2020-gpt3,dong-etal-2024-ICLsurvey}. 
By incorporating task-specific demonstrations directly into the input, ICL enables LLMs to adapt to specific tasks and generate more accurate outputs without requiring parameter updates. 
Recently, several efforts have been made on unveiling the underlying mechanisms of ICL \cite{zhou-etal-2024-mystery-survey,edelman2024evolution,park2025competition} and exploring methods to further boost the ICL performance \cite{wang2023large,rubin-etal-2022-learning,agarwal2024manyshot}.

\begin{figure}
    \centering
    \small
    \begin{subfigure}[b]{0.32\textwidth}
        \begin{tikzpicture}
    \begin{axis}[
        width=\textwidth,
        xlabel={},
        ylabel={Performance Change},
        ylabel style={rotate=0, anchor=south, yshift=-.5em},
        xtick={GSM8K, MATH},
        xticklabel style={
            align=center,
            font=\scriptsize
        },
        enlarge x limits=0.6, 
        bar width=6.5pt,      
        ybar,               
        axis lines*=left,
        extra y ticks={0},
        extra y tick style={
            grid=both,
            grid style={dashed,black},
            tick label style={opacity=0},
            major tick length=0pt
        },
        legend style={
            at={(0.4,-0.25)},
            anchor=north,
            legend columns=3,
            cells={anchor=west}, 
            font=\scriptsize, 
        },
        legend image code/.code={%
            \draw[#1, draw=none] (0cm,-0.1cm) rectangle (0.15cm,0.1cm);
        },
        ymin=-10, ymax=10,
        nodes near coords,
        nodes near coords style={
            font=\scriptsize,
            /pgf/number format/.cd,
            fixed,
            fixed zerofill,
            precision=1,
        },
        symbolic x coords={GSM8K, MATH}
    ]

    \addplot[fill=blue!30] coordinates {
        (GSM8K,10.0)
        (MATH,9.6)
    };
    \addlegendentry{Llama2-7B} 

    \addplot[fill=green!30] coordinates {
        (GSM8K,-1.8)
        (MATH,2.4)
    };
    \addlegendentry{Llama3.1-8B} 

    \addplot[fill=red!30] coordinates {
        (GSM8K,-6.0)
        (MATH,-1.2)
    };
    \addlegendentry{Llama-R1-8B} 

    \end{axis}
\end{tikzpicture}
        \vspace{-1.5em}
        \caption{}
        \label{fig:performance_comparison}
    \end{subfigure}
    \begin{subfigure}[b]{0.32\textwidth}
        \begin{tikzpicture}
    \begin{axis}[
        xmin=0, xmax=1,
        ymin=0, ymax=1,
        grid=both,
        width=1.1\textwidth,
        legend style={
            legend image post style={thick, line width=1pt},
            cells={anchor=west},         
            at={(0.03,0.97)},            
            anchor=north west,           
            draw=none,                   
            font=\tiny,                 
        },
        xlabel={Contextual Relevance},
        ylabel={Effectiveness Metric}
    ]

    \addplot [softblue, only marks, mark=*, mark size=2pt] coordinates {
        (0.21,0) (0.52,1) (0.59,0) (0.63,0)
        (0.66,1) (0.76,0) (0.87,1) (1.00,1)
    };
    \addlegendentry{Exact Match}

    \addplot [softred, only marks, mark=*, mark size=2pt] coordinates {
        (0.21,0.01) (0.52,0.15) (0.59,0.25) (0.63,0.53)
        (0.66,0.65) (0.76,0.75) (0.87,0.80) (1.00,1.00)
    };
    \addlegendentry{Loss Decrease}
    
    \end{axis}
\end{tikzpicture}
        \vspace{-1.5em}
        \caption{}
        \label{fig:performance_loss_compare}
    \end{subfigure}
    \begin{subfigure}[b]{0.32\textwidth}
        \includegraphics[width=\textwidth]{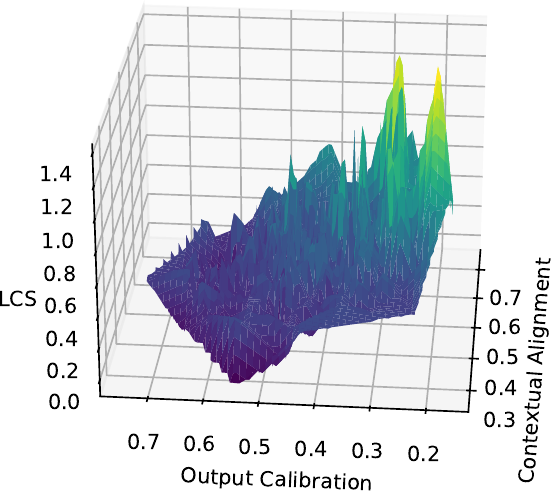}
        \vspace{-1.5em}
        \caption{}
        \label{fig:metric_3d}
    \end{subfigure}
    \caption{
        (a) Performance change of different models before and after applying ICL, where ICL exhibits varying effectiveness across different models on the same dataset.
        (b) Comparisons between metrics based on \textblue{exact match} and \textred{loss decrease}.
        Each dot denotes an example data of MATH using Llama3.1-8b with different demonstrations.
        Performance-based metrics with only binary values fail to quantify the varying contributions of different demonstrations to achieving correct results.
        In contrast, metrics based on loss decrease yield continuous values, enabling better reliability on measuring ICL effectiveness.
        (c) The impact of the contextual alignment and output calibration capabilities of the model on the \ourmetric metric. 
    }
    \label{fig:motivation}
\end{figure}
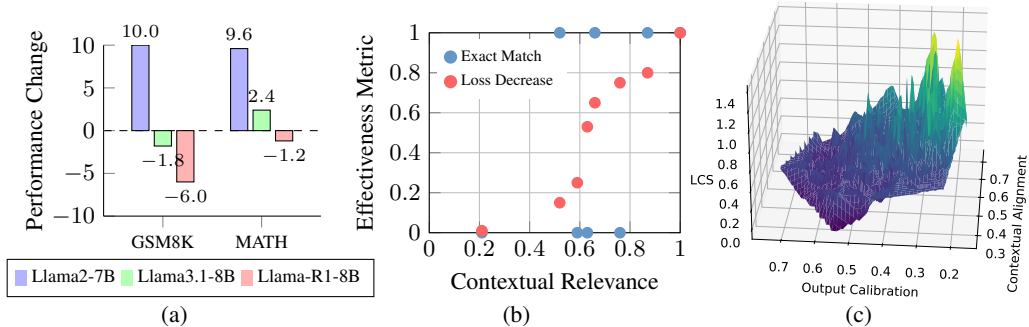

However, as illustrated in Figure~\ref{fig:performance_comparison}, even on the models with strong ICL capability like Llama3.1~\cite{grattafiori2024llama3}, ICL fails to enhance, and in some cases even harms, the performance \cite{deepseekai2025deepseekr1,huang2025explainable,zheng2025cursecotlimitationschainofthought}, showing different ICL effectivenes cross different models.
This observation raises a critical question: \textbf{How can practitioners reliably determine whether ICL is effective for a given model on a specific task}? 
This uncertainty poses practical challenges in real-world deployment of ICL: 
\begin{itemize}[leftmargin=*]
    \item For tasks with labeled data, practitioners often attempt to evaluate ICL effectiveness by observing performance changes after applying the selected demonstrations. However, this approach suffers from two critical limitations. \textit{(i) Low Reliability}: Performance fluctuations may stem from various factors like the quality of the insturction and selected demonstrations, making it difficult to isolate whether ICL itself is ineffective. \textit{(ii) Poor Attribution}: Disentangling the impact of individual factors requires costly repeated evaluations, hindering actionable analysis and insights.  
    \item For tasks without labeled data, there is no straightforward way to  assess whether adding demonstrations for ICL actually improves outcomes, leaving practitioners without guidance for improvments.
\end{itemize}

In light of these challenges, we propose a novel metric, named \textbf{L}earning-to-\textbf{C}ontext \textbf{S}lope (\textbf{\ourmetric}), which quantifies the ICL effectiveness by capturing the \textit{slope between the loss decrease by demonstrations (\textbf{learning gain}) and the demonstration relevance to the user input (\textbf{contextual relevance})}.
Specifically, \ourmetric is grounded in the perspective of the loss decrease in ICL \cite{wang2024loss,yang2024loss}.
For a given model and task, it evaluates how the learning gain varies with demonstrations of different contextual relevance.
This metric explicitly captures the two most important elements in \textit{in-context learning}: \textit{learning} and \textit{context} \cite{dong-etal-2024-ICLsurvey}. 
When ICL effectiveness is high, even demonstrations with low relevance can yield significant loss decrease. 
Conversely, when ICL effectiveness is low, only highly relevant demonstrations lead to noticeable learning gain.

Compared to performance-based measurement, \ourmetric offers the following advantages: 
\textit{(i) Higher Reliability}: As shown in Figure~\ref{fig:performance_loss_compare}, even when ICL fails to produce correct answers for user inputs, \ourmetric can still capture continous changes in model loss, providing a more reliable reflection of ICL effectiveness. 
\textit{(ii) Better Attribution}: \ourmetric is grounded in an intuitive mathematical formulation, enabling clearer analysis of how different factors influence ICL effectiveness. 
As shown in Figure~\ref{fig:metric_3d}, ICL tends to be ineffective when 1) the model fails to recognize the relevance of the demonstration to the input (\textit{i.e.}, the \textit{contextual alignment capability}), or 2) the model can independently verify the correctness of the output to the user input without adding demonstrations (\textit{i.e.}, the \textit{output calibration capability}).
\textit{(iii) Reduced Reliance on Labeled Evaluation Data}: We theoretically show that \ourmetric derived from synthetic data is consistently lower than that obtained from real data, and empirically identify a threshold value of \ourmetric indicative of effective ICL. 
Even in data-insufficient scenarios, \ourmetric can still offer actionable insights into ICL effectiveness.

Our contributions can be summarized as follows:
\begin{itemize}[leftmargin=*]
    \item 
        We propose a novel metric, namely Learning-to-Context Slope (\ourmetric), to measure the ICL effectiveness by capturing the two most important elements in ICL, including the learning gain and the contextual relevance of the demonstrations.
    \item 
        To validate the effectiveness of \ourmetric, we conduct extensive experiments on eight mainstream datasets covering mathematics, code, reasoning, and domain-specific tasks (\textit{e.g.}, finance and e-commerce). 
        The results validate a strong positive correlation between LCS and task performance improvements in scenarios where abundant labeled data enables reliable performance-based evaluation. 
        When labeled data exhibits inherent biases that distort performance-based metrics, \ourmetric consistently reflects true ICL effectiveness, underscoring its reliability. 
        Even without labeled data, LCS provides actionable insights into ICL effectiveness by leveraging synthetic data.
    \item 
        Further analysis identifies two key factors in LLMs that hinder ICL effectiveness: 
        1) weak contextual alignment capability to adapt inputs to task-specific demonstrations, and 
        2) strong output calibration capability to independently verify the correctness of outputs.
\end{itemize}

    \section{Proposed Metric: Learn-to-Context Slope}
        \label{sec:preliminary}
We introduce a novel metric, named Learn-to-Context Slope (LCS), to to measure the ICL effectiveness.
First, we interpret the ICL effectivess by measuring the loss decrease brought by using demonstrations based on the Bayesian model (\S \ref{subsec:effectiveness_measurement}).
Then, we present our LCS metric to measure the ICL effectiveness, based on which we discuss two main factors that influence the ICL effectiveness (\S \ref{subsec:metric}).
Further, we discuss the relationship between the metric using synthetic data and real data, aiding the application under the data-insufficient scenario (\S \ref{subsec:metric_without_label}). 

\subsection{Interpreting ICL Effectiveness via Loss Decrease}
    \label{subsec:effectiveness_measurement}

    Motivated by previous studies \cite{wang2024loss,yang2024loss}, the ICL effectiveness of a given predictive distribution $p$ with the parameter $\theta$ on a specific task with the task $\mathcal{C}=(Q,X,D)$ can be measured by the generation loss, \textit{i.e.,} negative log-likelihood:
    \begin{equation}
        \mathbb{L}_{\theta}(X | Q; D) = -\log{p(X|Q;D)},
    \end{equation}
    where $Q$ denotes the user input, $X$ represents the labeled output corresponding to $Q$, and $D$ denotes the demonstration, which are the random variables in the sampling spaces $\mathcal{X}$, $\mathcal{Y}$, and $\mathcal{D}$, respectively.
    Based on the Bayesian model, this loss can be obtained by:
    \begin{equation}
        \label{equ:loss_function_decomposed}
        \mathbb{L}_{\theta}(X | Q; D) = \mathbb{L}_{\theta}(X | Q) - (\log{p(D|Q;X)} - \log{p(D|Q)}),
    \end{equation}
    where $\mathbb{L}_{\theta}(X | Q)$ represents the loss of zero-shot generation, which is fixed given the model and task. 
    The proof of Equation~\ref{equ:loss_function_decomposed} is presented in Appendix~\ref{app:prove_of_loss_function}. It can be observed that only the second term, \textit{i.e.}, $\log{p(D|Q;X)} - \log{p(D|Q)}$, is relevant to the demonstrations, which is the loss decrease brought by the demonstrations. Intuitively, this term also measures the information of the user output $X$ that helps decide the demonstration $D$.

\subsection{Metric of ICL Effectiveness: \ourmetric}
    \label{subsec:metric}

    For simplicity, we denote the \textbf{Learning Gain} brought by the demonstrations as $I_p(X \rightarrow D | Q) = p(D|Q;X) - p(D|Q)$ to reflect the loss decrease. 
    To evaluate the overall ICL effectiveness of the given specific model and task, we propose to measure the effectiveness by evaluating how the learning gain varies with the demonstrations of different relevance. 
    The motivation is that even demonstrations with low relevance to the user question can still lead to significant learning gain for tasks and models where ICL is highly effective. 
    Conversely, when the ICL effectiveness is low, even the demonstration highly relevant to the user question can only lead to a small learning gain. 
    We measure the \textbf{Contextual Relevance} of the demonstration to the user question as $I_p(D \rightarrow X | Q) = p(X|Q;D) - p(X|Q)$. 
    The contextual relevance is quantified by how much information for inferring the output $X$ can be learned from the demonstration $D$ in the context. 
    We also compare the contextual relevance with other methods that measure the relevance to the user question of the demonstration in Appendix~\ref{app:similarity_measurement}.

    We have that the learning gain $I_p(X \rightarrow D | Q)$ and the contextual relevance $I_p(D \rightarrow X | Q)$ satisfy:

    \begin{theorem}
        $$I_p(X \rightarrow D | Q) = \frac{p(D|Q)}{p(X|Q)} I_p(D \rightarrow X | Q)$$
        \label{the:metric}
    \end{theorem}

    The proof of Theorem~\ref{the:metric} is presented in Appendix~\ref{app:prove_of_metric}.
    According to the theorem, learning gain and contextual relevance are positively correlated with a certain slope.
    A larger slope indicates a greater decrease in loss when increasing the information relevant to the user question of the demonstrations, thereby making ICL more effective.

    In practice, let $\hat{p}$ represent the empirical probability distribution and $C=\{(q_i,x_i,d_i)\}_n$ be the sampling on $\mathcal{C}$, we calculate the slope of Theorem~\ref{the:metric} ($r_{\hat{p}}$) on $C$ with the least squares method \cite{wang2018generalized}:
    \begin{equation}
        \begin{split}
            & r_{\hat{p}} = \frac{\sum_{i=1}^{n}(t_i-\bar{t})(s_i-\bar{s})}{\sum_{i=1}^{n}(t_i-\bar{t})^2}, \texttt{where} \\
            & s_i = I_{\hat{p}}(d_i \rightarrow x_i | q_i), t_i = I_{\hat{p}}(x_i \rightarrow d_i | q_i) \\
            & \bar{s} = \frac{1}{n}\sum_{i=1}^{n}s_i, \bar{t} = \frac{1}{n}\sum_{i=1}^{n}t_i.
        \end{split}
    \end{equation}
    We use \bm{$r_{\hat{p}}$} as the metric to measure the ICL effectiveness, which we call the \textbf{Learning-to-Relevance Ratio (\ourmetric)}.
    Although $r_p=\frac{p(D|Q)}{p(X|Q)}$, considering that $\hat{p}$ has the error compared with $p$, $r_{\hat{p}} \neq \frac{\hat{p}(D|Q)}{\hat{p}(X|Q)}$.
    In Appendix~\ref{app:error_of_metric}, we discuss the impact of error and prove that the impact of the error on $r_{\hat{p}}$ is less than $\frac{\hat{p}(D|Q)}{\hat{p}(X|Q)}$. 
    We discuss how to calculate our metric in detail in Appendix~\ref{app:metric_calculation}.

    Based on Theorem~\ref{the:metric}, it can be observed that there are two main factors influencing the ICL effectiveness: \textit{\textbf{the contextual alignment capability}} that learn the question-relevant information from the demonstrations ($\hat{p}(D|Q)$), and \textbf{\textit{the output calibration capability}} that verify the correctness of the output to the given input ($\hat{p}(X|Q)$). 
    Therefore, given a specific model and task, the reasons for poor ICL effectiveness can be attributed to two aspects:  
    \textit{(i) Low Contextual Alignment Ability}: The model fails to adequately comprehend the task-relevent information in provided demonstrations.
    \textit{(ii) High Output Calibration Capability}: The model possesses a strong inherent ability to verify the input consistency to the given output.
    We further discuss the meaning of the contextual alignment capability and the output calibration ability in detail in Appendix~\ref{app:effective_factor}.

\subsection{ICL Effectiveness without Labeling}
    \label{subsec:metric_without_label}

    Since the calculation of \ourmetric in \S~\ref{subsec:metric} relies on labeled data, its application to new tasks in data-insufficient scenarios is limited.
    Prior works have shown that the resource requirements for obtaining task questions are lower than those for obtaining the labels \cite{shen2018ordered,tan-etal-2024-large}.
    Therefore, in this section, we discuss the relationship of \ourmetric with the synthetic data and real data only with the labeled input, which satisfies that:

    \begin{theorem}
        Let $\hat{X} = \argmax_{X \sim \mathcal{X}} \hat{p}(X | Q), X^* = \argmax_{X \sim \mathcal{X}} p(X | Q)$.
        Suppose $\hat{D}, D^*$ satisfy that $\hat{p}(X | Q; \hat{D}) \leq \hat{p}(X | Q; D^*), \forall X \sim \mathcal{X},Q \sim \mathcal{Q}$.
        It can be concluded that:
        $$\frac{\hat{p}(\hat{D}|Q)}{\hat{p}(\hat{X}|Q)} \leq \frac{\hat{p}(D^*|Q)}{\hat{p}(X^*|Q)}$$
        \label{the:metric_without_labeling}
    \end{theorem}
    We provide the proof of Theorem~\ref{the:metric_without_labeling} in Appendix~\ref{app:prove_of_metric_without_labeling}. 
    About Theorem~\ref{the:metric_without_labeling}, \(\hat{X}\) and \(X^*\) can be seen as the predicted and real output, respectively. 
    The conditions in the theorem can be interpreted as stating that, for any given data, real demonstrations always better assist in correctly answering the question compared to synthetic demonstrations.
    Based on Theorem~\ref{the:metric_without_labeling}, we can observe that the $r_{\hat{p}}$ fitted with synthetic data is consistently smaller than using real data. 
    Consequently, while fitting synthetic data can reflect the ICL effectiveness to some extent, the magnitude of the effectiveness derived is lower than the real effectiveness.

    \section{Experiment}
        \label{sec:experiment}
In this section, we empirically investigate three research questions about the ICL effectiveness:
\textbf{\textsc{RQ1}}. How to reliably evaluate the ICL effectiveness?
\textbf{\textsc{RQ2}}. How does different factors influence the ICL effectiveness?
\textbf{\textsc{RQ3}}. Can synthetic data accurately reflect the ICL effectiveness?

\subsection{Experiment Setup}
    \label{subsec:experiment_setup}

    \paragraph{Dataset}
        We conduct experiments on four mainstream tasks: math (GSM8K~\cite{cobbe2021gsm8k}, MATH~\cite{hendrycks2021math}), code (HumanEval~\cite{chen2021humaneval}, MBPP~\cite{austin2021mbpp}), reason (ARC-Challenge~\cite{yadav-etal-2019-arc}, MMLU-Pro~\cite{wang2024mmlupro}), and domain-specific (FinQA~\cite{chen2021finqa}, Amazon Review~\cite{ni-etal-2019-amazon}).
        We introduce the above dataset, as well as the split of the demonstrations and the test data in Appendix~\ref{app:benchmark}.
        For datasets of math, reasoning, and domain-specific, we use Exact Match (EM)~\cite{cobbe2021gsm8k} as the evaluation metric.
        For the datasets of code, we use Pass@1~\cite{chen2021humaneval} as the metric.
        We suppose that the demonstrations of the aforementioned datasets can genuinely reflect the ICL effectiveness through performance improvements, thereby validating the effectiveness of \ourmetric. 
        In \S\ref{subsubsec:effectiveness_performance}, we present that \ourmetric can still reflect the ICL effectiveness even when the provided demonstrations do not lead to performance improvements.

    \paragraph{Model}
        We conduct experiments on three mainstream LLMs: Llama2-7b~\cite{touvron2023llama2}, Llama3.1-8b~\cite{grattafiori2024llama3}, and DeepSeek-R1-Distill-Llama-8b (Llama-R1-8b) \cite{deepseekai2025deepseekr1}, which cover different ICL capabilities to fully evaluate our metric.
        We also conduct experiments on models of other scales and series in Appendix~\ref{app:experiment_model_type}, further validating the effectiveness of \ourmetric.

    \paragraph{Implementation Details}
        We evaluate the performance on all datasets under 0-shot and 1-shot settings, using BM25 to select the demonstrations for each user input.
        We also discuss the performance and ICL effectiveness under different shots in \S\ref{subsubsec:sample_scle}.
        Following \citep{deepseekai2025deepseekr1}, we set the maximum generation length to $32768$.
        Our experiments are conducted on a single A100-80G, with an average computation time of approximately $20$ minutes on each dataset and model.

\subsection{RQ1. How to Reliably Evaluate the Effectiveness of ICL?}
    First, we discuss that \ourmetric can accurately reflect the effectiveness of ICL. 
    Subsequently, we provide experimental evidence demonstrating that performance improvement is insufficient for accurately reflecting the ICL effectiveness.
    In addition, we present that \ourmetric can reflect the performance improvement brought by ICL to a certain extent.

    \subsubsection{\ourmetric Reliably Reflects the ICL Effectiveness}
        \label{subsec:main_experiment}
    
        \begin{table*}[t]
            \centering
            \small
            \caption{
                Performance and fitted lines across different models and datasets.
                ARC-C denotes ARC-Challenge, and Amazon denotes Amazon Review.
                $\Delta$ denotes the performance change of 1-shot compared to 0-shot, where performance gains $< 1.0$ are marked in \textred{red}.  
                $r_{\hat{p}}x + b$ represents the fitted line discussed in \S\ref{subsec:metric}, where $\ourmetric (r_{\hat{p}}) < 0.2$ are highlighted in \textred{red}. 
                Appendix~\ref{app:experiment_result} shows the detailed performance and figures.
            }
            \begin{tabular}{l|cl|cl|cl}
    \toprule
    \multirow{2}{*}{\textbf{Dataset}} & \multicolumn{2}{c|}{\textbf{Llama2-7b}} & \multicolumn{2}{c|}{\textbf{Llama3.1-8b}} & \multicolumn{2}{c}{\textbf{Llama-R1-8b}} \\
     & $\Delta$ & $r_{\hat{p}}x + b$ & $\Delta$ & $r_{\hat{p}}x + b$ & $\Delta$ & $r_{\hat{p}}x + b$ \\
    \midrule
    GSM8K & $+10.0$ & $0.32 x + 0.02$ & \textred{{$-1.8$}} & \textred{{$0.07 x + 0.12$}} & \textred{{$-6.0$}} & \textred{{$0.05 x - 0.03$}} \\
    MATH & $+9.6$ & $1.03 x + 0.03$ & $+2.4$ & $0.34 x - 0.00$ & \textred{{$-1.2$}} & \textred{{$0.09 x - 0.03$}} \\
    HumanEval & \textred{{$-0.6$}} & \textred{{$0.07 x + 0.00$}} & \textred{{$-2.5$}} & \textred{$0.10 x + 0.02$} & \textred{{$-3.0$}} & \textred{{$-0.11 x - 0.02$}} \\
    MBPP & \textred{{$-0.5$}} & \textred{{$0.05 x + 0.02$}} & \textred{{$+0.8$}} & \textred{$0.15 x + 0.07$} & \textred{{$-6.4$}} & \textred{{$0.07 x + 0.00$}} \\
    ARC-C & $+11.6$ & $0.74 x + 0.10$ & \textred{{$-1.9$}} & \textred{{$-0.54 x - 0.08$}} & \textred{{$-0.3$}} & \textred{{$0.08 x - 0.05$}} \\
    MMLU-Pro & $+5.5$ & $0.64 x + 0.11$ & $+2.6$ & $0.52 x - 0.01$ & \textred{{$-5.5$}} & \textred{{$-0.04 x - 0.01$}} \\
    FinQA & $+7.3$ & $0.63 x + 0.02$ & $+4.9$ & $0.82 x - 0.06$ & \textred{{$-1.8$}} & \textred{{$0.04 x - 0.04$}} \\
    Amazon & \textred{{$+0.5$}} & \textred{{$0.07 x - 0.06$}} & $+5.0$ & $0.94 x - 0.09$ & $+11.8$ & $0.37 x - 0.11$ \\
    \bottomrule
\end{tabular}

            \label{tab:main_experiment}
        \end{table*}

        Accodring to the main experimental results shown in Table~\ref{tab:main_experiment}, there are several notable observations:

        \textbf{\textit{The ICL effectiveness is independent of dataset difficulty.}}
            For Llama2-7b, ICL is effective on the MATH dataset but fails on the easier Amazon Review dataset.
            Conversely, for Llama-R1-8b, ICL is ineffective on MATH but performs well on Amazon Review. 
            This discrepancy arises because, for more difficult datasets, the model struggles to comprehend the relationships between demonstrations, answers, and user questions, leading to a decline in both the ICL abality and the answer verification ability.
            Consequently, it is uncertain whether \ourmetric increases or decreases on more difficult datasets.
    
        \textbf{\textit{The ICL effectiveness is independent of model capability.}}
            In Amazon Review, Llama-R1-8b demonstrates a significant improvement with ICL, whereas the less capable Llama2-7b does not exhibit a noticeable performance improvement. 
            This discrepancy arises because, as the model capability increases, both the contextual alignment capability and the output calibration capability increase simultaneously, making it uncertain whether \ourmetric rises or falls.

        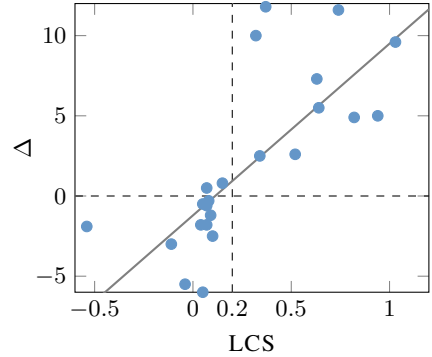
\begin{wrapfigure}{r}{0.45\textwidth}
        \begin{minipage}[t]{0.45\textwidth}
            \centering
            \small
            \begin{tikzpicture}
    \small
    \begin{axis}[
        xlabel={\ourmetric},
        ylabel={$\Delta$},
        ylabel style={rotate=0, anchor=south, yshift=-.5em},
        legend style={at={(0.5,-0.2)}, anchor=north, legend columns=2},
        xmajorgrids=false,
        ymajorgrids=false,
        extra x ticks={0.2},
        extra x tick style={
            grid=both,
            grid style={dashed,black},
            tick label style={
                /pgf/number format/fixed,
                /pgf/number format/precision=3
            },
            major tick length=0pt
        },
        extra y ticks={0},
        extra y tick style={
            grid=both,
            grid style={dashed,black},
            tick label style={opacity=0},
            major tick length=0pt
        },
        xmin=-0.6, xmax=1.2,
        ymin=-6, ymax=12,
        width=\textwidth,
        every axis legend/.append style={nodes={anchor=west}},
        xticklabel style={
            /pgf/number format/fixed,
            /pgf/number format/precision=3
        },
        yticklabel style={
            /pgf/number format/fixed,
            /pgf/number format/precision=1
        }
    ]

    \addplot[only marks, color=softblue, mark size=2pt, mark=*] coordinates {
        (0.32, 10.0)
        (1.03, 9.6)
        (0.07, -0.6)
        (0.05, -0.5)
        (0.74, 11.6)
        (0.64, 5.5)
        (0.63, 7.3)
        (0.07, 0.5)
        (0.07, -1.8)
        (0.34, 2.5)
        (0.10, -2.5)
        (0.15, 0.8)
        (-0.54, -1.9)
        (0.52, 2.6)
        (0.82, 4.9)
        (0.94, 5.0)
        (0.05, -6.0)
        (0.09, -1.2)
        (-0.11, -3.0)
        (0.07, -6.4)
        (0.08, -0.3)
        (-0.04, -5.5)
        (0.04, -1.8)
        (0.37, 11.8)
    };

    \addplot [domain=-0.6:1.2, samples=100, color=gray, thick] {-1.2093 + 10.711*x};



    \end{axis}
\end{tikzpicture}
            \caption{
                The performance improvement $\Delta$ brought by ICL (y-axis) with different $\ourmetric$ (x-axis) on different models and datasets.
                The solid line in the graph represents the fitted line for all data points.
                The two dashed lines represent $\ourmetric = 0.2$ and $\Delta = 0$, respectively.
            }
            \vspace{-4mm}
            \label{fig:performance_slope}
        \end{minipage}
        \end{wrapfigure}
        \textbf{\textit{The performance improvement brought by ICL is positively related to \ourmetric.}} 
        To evaluate whether \ourmetric effectively reflects the efficacy of ICL, we analyze the performance improvement with different \ourmetric, as illustrated in Figure~\ref{fig:performance_slope}. 
        The Pearson correlation coefficient between \ourmetric and the performance change is $0.737$, demonstrating a positive correlation.
        A high \ourmetric suggests that the model achieves higher learning gain as the contextual relevance increases, demonstrating that LLMs learn how to solve the task from the demonstrations, thereby improving the performance. 
        In contrast, a low \ourmetric indicates that the learning gain from the demonstrations remains relatively constant regardless of the contextual relevance, implying limited learning from the demonstrations.
        Specifically, based on the Figure~\ref{fig:performance_slope}, we can use $\ourmetric = 0.2$ as a empirical threshold, since when \( \ourmetric \leq 0.2 \), the corresponding performance gain is minimal or negative, suggesting that ICL is less effective in the given task and model.

        \subsubsection{The Performance Change Cannot Reflect the ICL Effectiveness}
            \label{subsubsec:effectiveness_performance}
        
            \begin{figure}
                \centering
                \small
                \begin{subfigure}[b]{0.4\textwidth}
                  \includegraphics[width=\textwidth]{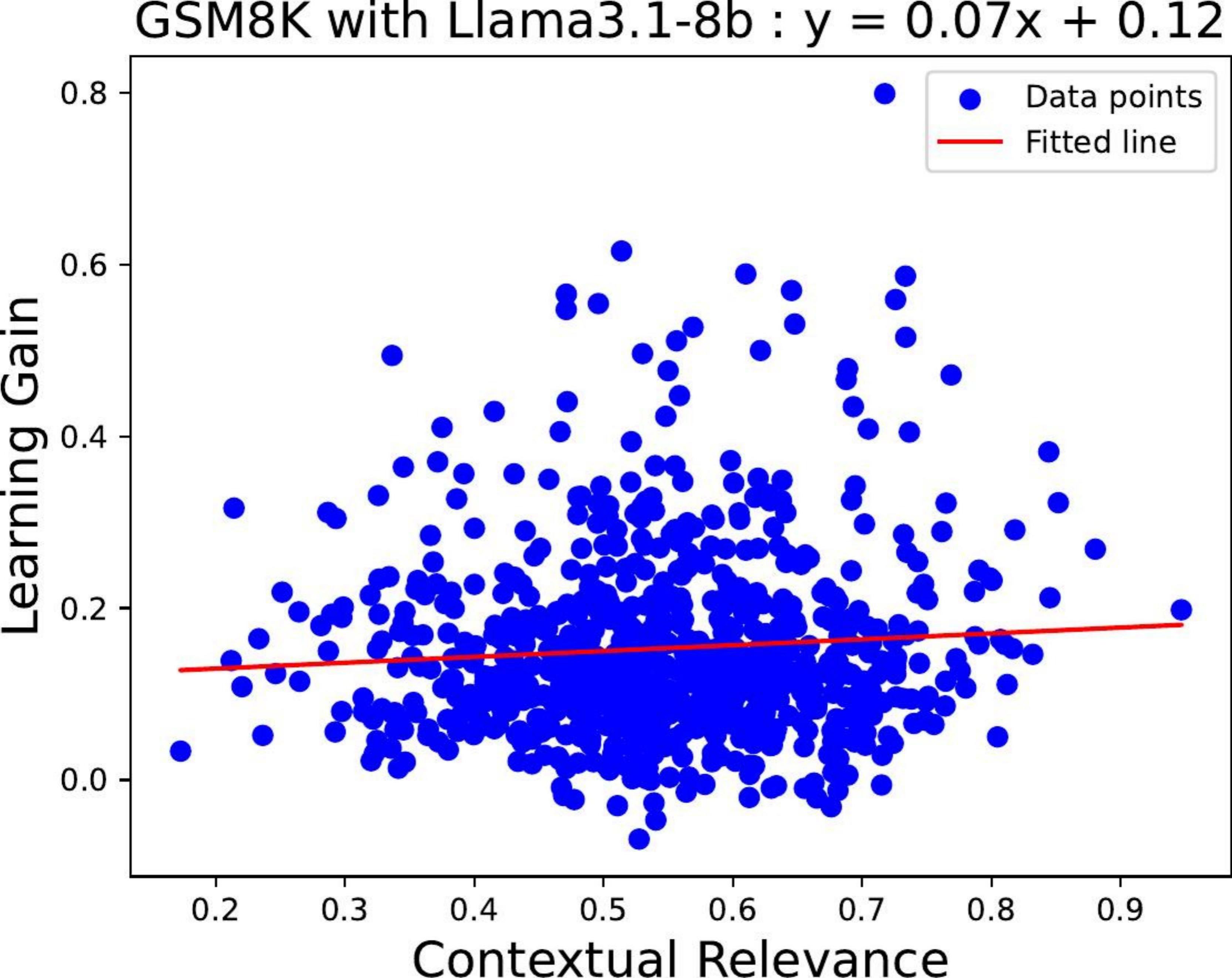}
                  \vspace{-1.5em}
                  \caption{All Cases, $\Delta=-1.8$, \ourmetric$=0.07$.}
                \end{subfigure}
                \hspace{3em}
                \begin{subfigure}[b]{0.4\textwidth}
                  \includegraphics[width=\textwidth]{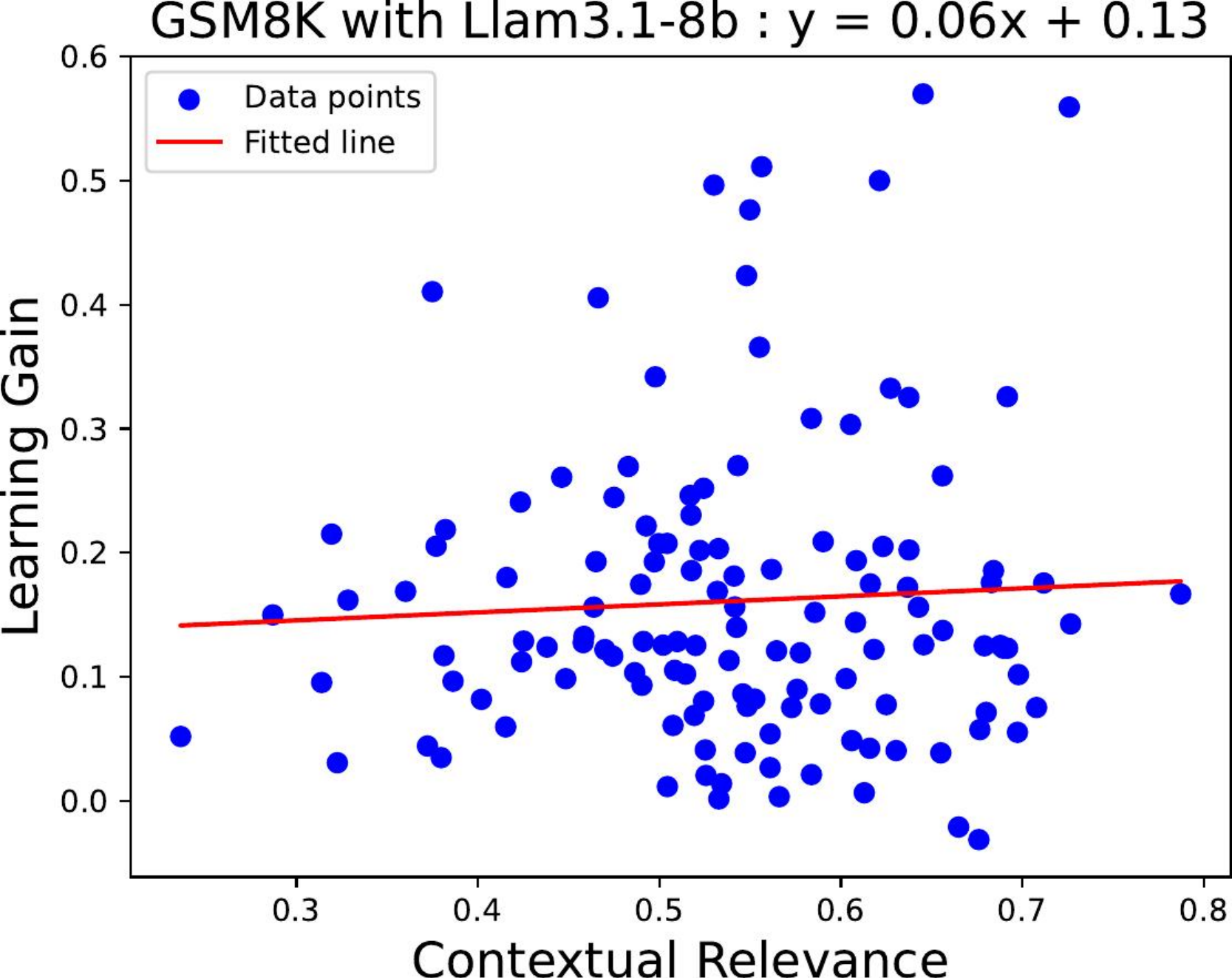}
                  \vspace{-1.5em}
                  \caption{Bad Cases, $\Delta=+0.0$, \ourmetric$=0.06$.}
                \end{subfigure}
    
                \begin{subfigure}[b]{0.4\textwidth}
                  \includegraphics[width=\textwidth]{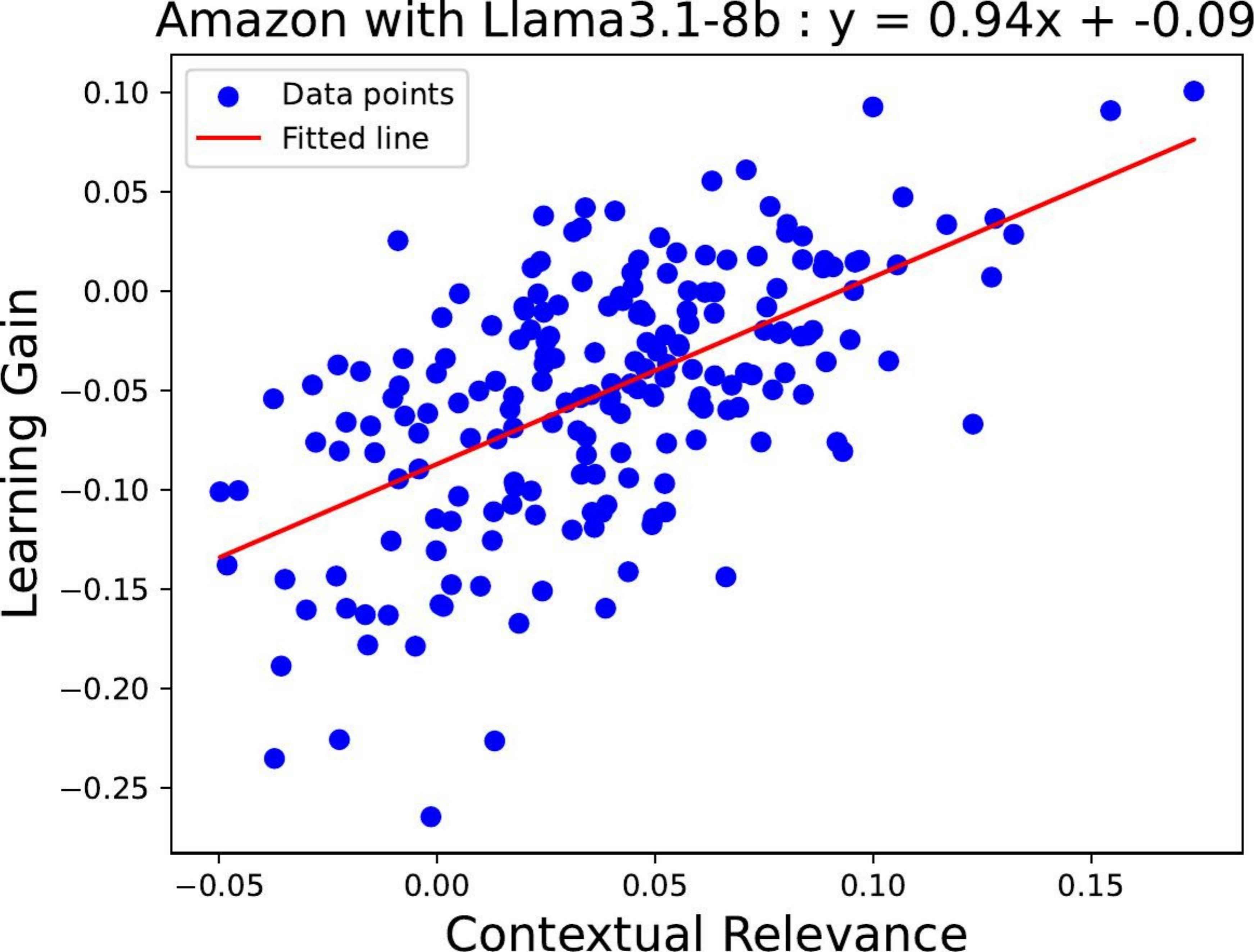}
                  \vspace{-1.5em}
                  \caption{All Cases, $\Delta=+5.0$, \ourmetric$=0.94$.}
                \end{subfigure}
                \hspace{3em}
                \begin{subfigure}[b]{0.4\textwidth}
                  \includegraphics[width=\textwidth]{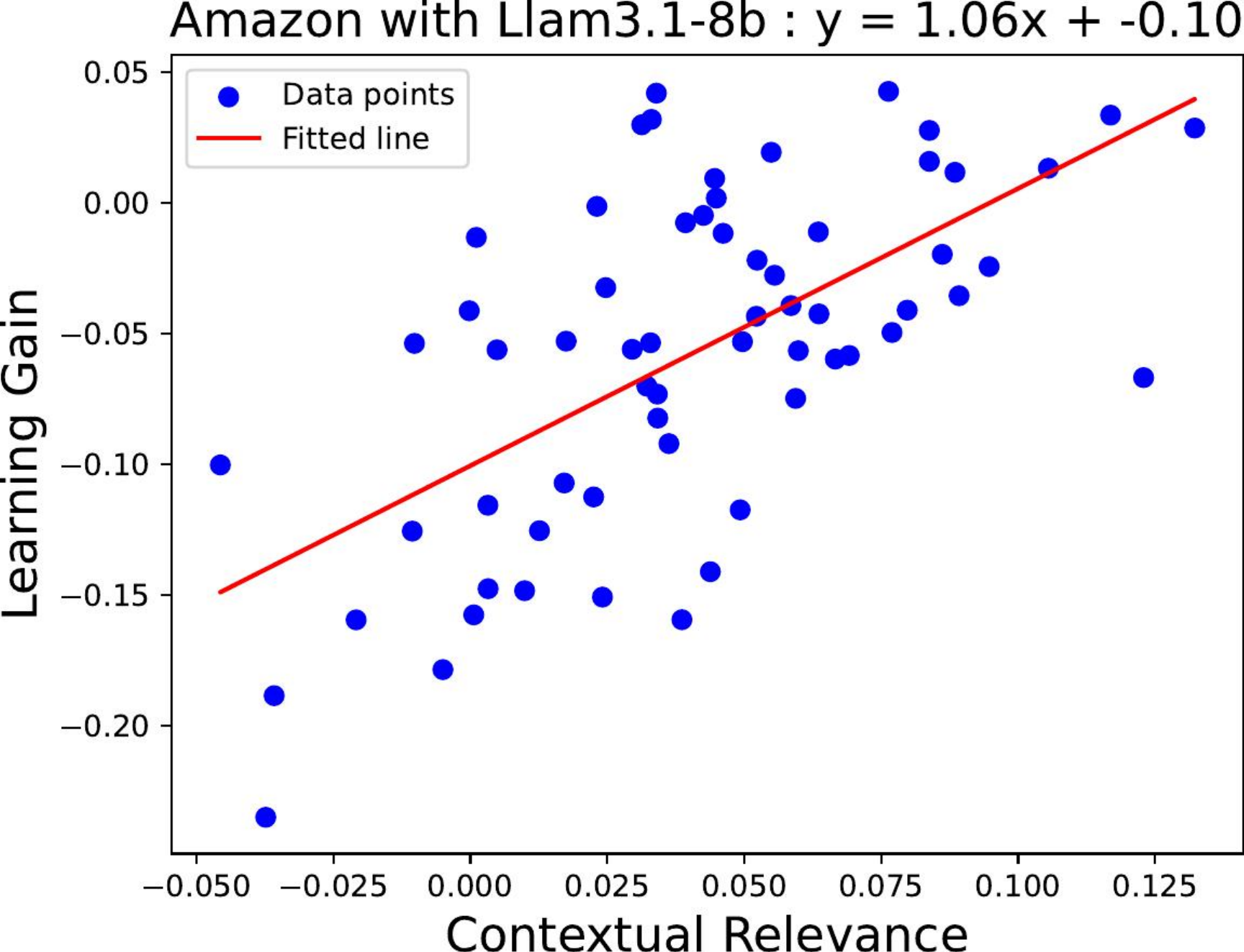}
                  \vspace{-1.5em}
                  \caption{Bad Cases, $\Delta=+0.0$, \ourmetric$=1.06$.}
                \end{subfigure}
                \hfill
                \caption{
                    The experimental results of using Llama3.1-8b on GSM8K and Amazon under the full set and the bad cases of ICL.
                    $\Delta$ denotes the performance change of ICL compared with zero-shot.
                }
                \label{fig:metric_under_bad_cases}
            \end{figure}
    
            In \S\ref{subsec:experiment_setup}, we assume that whether performance improves or not can genuinely reflect the ICL effectiveness on all experiment datasets.  
            However, in practical applications, the quality of demonstrations or instructions can impact performance, causing no performance improvement even for models and tasks where ICL is effective.
            To demonstrate that \ourmetric can still reflect the ICL effectiveness even when performance does not improve, we plot $r_{\hat{p}}$ on the bad cases after using ICL, as shown in Figure~\ref{fig:metric_under_bad_cases}.  
            It can be observed that:
            \textit{(i)} Even on data where ICL does not improve performance, \ourmetric still reveals the ICL effectiveness, proving the higher reliability of our metric compared with the performance-based metric.
            \textit{(ii)} \ourmetric is higher reliable, unlike performance which is susceptible to the factors like the instruction, as it directly evaluates $p(X|Y)$ by using $Y$ as input and $X$ as output without relying on instructions (Appendix~\ref{app:metric_calculation}), thus providing a more faithful reflection of the ICL effectiveness.

    \subsubsection{The Learning Gain is a Good Metric for Demonstration Selection}
        \begin{wraptable}{R}{0.55\textwidth}
            \centering
            \small
            \vspace{-4mm}
            \caption{
                The performance of ICL with different demonstration selection methods using Llama3.1-8b.
                The best performance of each setting is marked in \textbf{bold}.
            }
            \setlength{\tabcolsep}{1.5mm}{\begin{tabular}{l|cccc}
    \toprule
    \textbf{Method} & \textbf{MATH} & \textbf{MMLU-Pro} & \textbf{FinQA} & \textbf{Amazon} \\
    \midrule
    Zero-Shot & $48.4$ & $50.4$ & $49.7$ & $63.5$ \\
    BM25~\cite{robertson-etal-2009-bm25} & $50.8$ & $53.0$ & $54.6$ & $68.5$ \\
    GTR~\cite{luo2023dricl} & $50.8$ & $53.5$ & $55.0$ & $68.5$ \\
    IDS~\cite{qin-etal-2024-ids} & $50.4$ & $52.4$ & $54.6$ & $68.0$ \\
    Ours & \bm{$51.2$} & \bm{$54.3$} & \bm{$55.1$} & \bm{$70.0$} \\
    \bottomrule
\end{tabular}}

            \vspace{-1em}
            \label{tab:demonstration_selection}
        \end{wraptable}
    
        Enhancing ICL performance has been a topic of significant interest. 
        Although this paper does not primarily focus on improving ICL performance, the discussions in \S\ref{subsec:main_experiment} reveal several potential avenues for improvement.
        We observe that while there is a general positive correlation between the loss decrease and the information learned from demonstrations, there also exist cases where demonstrations with rich information yield low loss decrease.
        To address this, we propose a method that first generates a preliminary answer $\hat{X}$ for the user question and then selects the most appropriate demonstrations based on learning gain.
        To validate the effectiveness of this method, we conduct experiments using Llama3.1-8b on the MATH and Amazon datasets. 
        As shown in Table~\ref{tab:demonstration_selection}, our method outperforms other demonstration selection baselines, demonstrating the effectiveness of the method based on learning gain.


\subsection{RQ2. How Does Different Factors Influence the ICL Effectiveness?}

    \subsubsection{The Main Factors that Influence the ICL Effectiveness}
        \begin{wrapfigure}{r}{0.5\textwidth}
        \vspace{-4mm}
            \centering
            \small
            \begin{tikzpicture}
    \begin{axis}[
        width=0.55\textwidth,
        height=4cm,
        xlabel={},
        ylabel={},
        symbolic x coords={ICL, Verify, Metric},
        xticklabels={Contextual Alignment, Output Calibration, \ourmetric},
        xtick=data, 
        xticklabel style={align=center, font=\scriptsize}, 
        enlarge x limits=0.22, 
        ybar,                  
        bar width=9pt,        
        legend style={
            at={(0.5,-0.3)}, 
            anchor=north,
            legend columns=2,
            font=\small      
        },
        ymin=0, ymax=1,
        nodes near coords, 
        nodes near coords style={
            font=\tiny, 
            /pgf/number format/.cd,
            fixed,         
            fixed zerofill,
            precision=2,   
        },
    ]

    \addplot[fill=blue!30] coordinates {
        (ICL, 0.44) (Verify, 0.35) (Metric, 0.32) 
    };
    \addlegendentry{GSM8K, Llama2}

    \addplot[fill=red!30] coordinates {
        (ICL, 0.13) (Verify, 0.37) (Metric, 0.07) 
    };
    \addlegendentry{GSM8K, Llama3.1} 

    \addplot[fill=green!30] coordinates {
        (ICL, 0.57) (Verify, 0.65) (Metric, 0.07) 
    };
    \addlegendentry{Amazon, Llama2} 

    \addplot[fill=yellow!30] coordinates {
        (ICL, 0.72) (Verify, 0.66) (Metric, 0.94) 
    };
    \addlegendentry{Amazon, Llama3.1}

    \end{axis}
\end{tikzpicture}
            \caption{
                The results of the contextual alignment capability ($\hat{p}(D|Q)$), the output calibration capability ($\hat{p}(X|Q)$) and \ourmetric of Llama2-7b and Llama3.1-8b on GSM8K and Amazon Review.
                $\hat{p}(D|Q)$ and $\hat{p}(X|Q)$ are calculated as the average value on all test data.
            }
            \label{fig:factor_analysis}
            \vspace{-1mm}
        \end{wrapfigure}
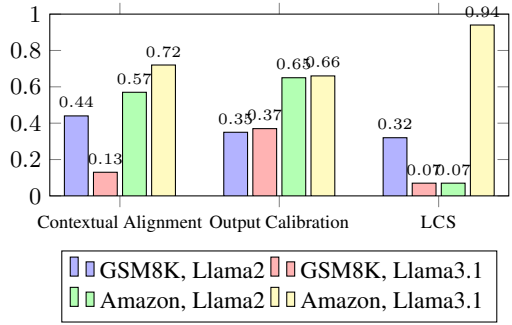
    
        In \S\ref{subsec:metric}, we discuss the main factors influencing the ICL effectiveness, including the contextual alignment capability ($\hat{p}(D|Q)$) and the output calibration capability ($\hat{p}(X|Q)$).
        In this section, we conduct experiments to analyze these conclusions further.
        We calculate the average values of $\hat{p}(D|Q)$ and $\hat{p}(X|Q)$ with Llama2-7b and Llama3.1-8b on GSM8K and Amazon Review, as shown in Figure~\ref{fig:factor_analysis}.
        From the figure, we can observe the following:
        \textit{(i)} The results of Llama2-7b on Amazon Review indicate that the model is unable to effectively learn the information relevant to the user input from the provided demonstration $D$, \textit{i.e.}, the contextual alignment capability is low, which leads to poor ICL effectiveness;
        \textit{(ii)} The results of Llama3.1-8b on GSM8K show that although the ICL ability of the model is high, the model can accurately assess the relationship between input and output, \textit{i.e.}, the output calibration capability, also diminishes the ICL effectiveness;
        \textit{(iii)} \ourmetric is not equal to $\frac{\hat{p}(D|Q)}{\hat{p}(X|Q)}$, due to the error between $p$ and $\hat{p}$, as discussed in detail in Appendix~\ref{app:error_of_metric}.

    \subsubsection{Shot Number Does Not Affect the ICL Effectiveness Obviously}  
        \label{subsubsec:sample_scle}

        \begin{table}[t]
            \centering
            \small
            \caption{
                The performance and the fitted lines of Llama2-7b and Llama3.1-8b on MATH and Amazon Review with different shots.
                $\Delta$ denotes the performance change compared with 0-shot.
            }
            \begin{tabular}{lc|cl|cl}
    \toprule
    \multirow{2}{*}{\textbf{Dataset}} & \multirow{2}{*}{\textbf{Shot}} & \multicolumn{2}{c|}{\textbf{Llama2-7b}} & \multicolumn{2}{c}{\textbf{Llama3.1-8b}} \\
     & & \bm{$\Delta$} & \bm{$r_{\hat{p}}x + b$} & \bm{$\Delta$} & \bm{$r_{\hat{p}}x + b$} \\
    \midrule
    \multirow{4}{*}{MATH} & $1$ & $+9.6$ & $1.03x+0.03$ & $+2.4$ & $0.34x-0.00$ \\
     & $2$ & $+10.0$ & $1.11x+0.04$ & $+2.0$ & $0.30x+0.01$ \\
     & $3$ & $+10.4$ & $1.01x+0.05$ & $+2.2$ & $0.34x+0.01$ \\
     & $4$ & $+10.2$ & $0.94x+0.04$ & $+1.8$ & $0.32x+0.01$ \\
    \midrule
    \multirow{4}{*}{Amazon} & $1$ & $+0.5$ & $0.07x-0.06$ & $+5.0$ & $0.94 x - 0.09$ \\
     & $2$ & $-0.5$ & $0.05x-0.08$ & $+5.0$ & $0.98 x - 0.08$ \\
     & $3$ & $+0.0$ & $0.07x-0.08$ & $+5.5$ & $0.95 x - 0.09$ \\
     & $4$ & $+0.5$ & $0.05x-0.07$ & $+6.0$ & $0.98x - 0.08$ \\
    \bottomrule
\end{tabular}

            \label{tab:sample_scale}
        \end{table}

        To observe the differences in the ICL effectiveness under different shot numbers, we conduct experiments under different shot number. 
        Since Theorem~\ref{the:metric} can only calculate the influence of a single demonstration, we divide the k-shot into k data points to calculate \ourmetric. 
        The experimental results are shown in Table~\ref{tab:sample_scale}. 
        From the table, it can be observed that although increasing the number of shots enhances performance, it has little impact on the computation of \ourmetric. 
        This is because \ourmetric is determined by the model and task, and is irrelevant to the shot number.
        Therefore, increasing the number of shots merely decreases the error bewtween $p$ and $\hat{p}$, causing less changes of \ourmetric.

    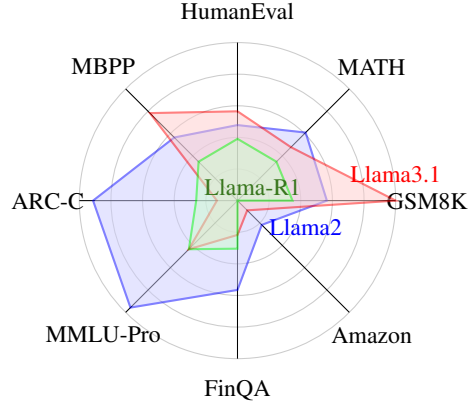
\begin{wrapfigure}{r}{0.45\textwidth}
        \begin{minipage}[b]{0.43\textwidth}
            \vspace{-4.5mm}
            \centering
            \small
            \begin{tikzpicture}[
  font=\small,
  x=0.035\textwidth,
  y=0.035\textwidth
]
  \useasboundingbox (-12, -12) rectangle (12, 12);

  \coordinate (origin) at (0,0);

  \def\n{8}                 
  \def\step{360/\n}         
  \def\labels{{"GSM8K","MATH","HumanEval","MBPP",
               "ARC-C","MMLU-Pro","FinQA","Amazon"}} 

  \def\ModelA{{5.65, 6.09, 4.78, 5.65, 9.13, 9.57, 5.65, 2.17}}
  \def\ModelB{{10.00, 4.78, 5.65, 7.83, 1.30, 4.35, 2.17, 0.87}}
  \def\ModelC{{3.48, 3.48, 3.91, 3.48, 2.61, 4.35, 3.04, 0.00}}

  \def\maxRadius{10}      
  \def\labelRadius{12}    

  \foreach \r in {2,4,6,8,10} {
      \draw[gray!40] (origin) circle (\r);
  }

  \foreach \i in {1,...,\n} {
      \pgfmathtruncatemacro{\deg}{(\i-1)*\step}
      \draw (origin) -- (\deg:\maxRadius);
  }

  \foreach \i in {1,...,\n} {
      \pgfmathtruncatemacro{\deg}{(\i-1)*\step}
      \pgfmathparse{\labels[\i-1]}
      \let\dimName\pgfmathresult
      \node at (\deg:\labelRadius) {\dimName};
  }

  \foreach \i in {1,...,\n} {
      \pgfmathtruncatemacro{\j}{\i-1}

      \pgfmathparse{\ModelA[\j]}\let\valA\pgfmathresult
      \coordinate (A\i) at (\j*\step:\valA);

      \pgfmathparse{\ModelB[\j]}\let\valB\pgfmathresult
      \coordinate (B\i) at (\j*\step:\valB);

      \pgfmathparse{\ModelC[\j]}\let\valC\pgfmathresult
      \coordinate (C\i) at (\j*\step:\valC);
  }

  \draw[fill=blue!20, draw=blue, thick, opacity=0.5]
       (A1) \foreach \i in {2,...,\n}{--(A\i)} -- cycle;

  \draw[fill=red!20, draw=red, thick, opacity=0.5]
       (B1) \foreach \i in {2,...,\n}{--(B\i)} -- cycle;

  \draw[fill=green!20, draw=green, thick, opacity=0.5]
       (C1) \foreach \i in {2,...,\n}{--(C\i)} -- cycle;

  \node[blue, xshift=-8, yshift=-10] at (A1) {Llama2};
  \node[red, xshift=0,  yshift= 10] at (B1) {Llama3.1};
  \node[softgreen, xshift=-15, yshift=5] at (C1) {Llama-R1};

\end{tikzpicture}
            \caption{
                The intercept of the fitted line on each dataset and each model.
                We also compare the intercepts of Llama3.1 under different scales in Appendix~\ref{app:intercept_scales}.
            }
            \label{fig:information_gain}
            \vspace{-4.5mm}
        \end{minipage}
    \end{wrapfigure}

    \subsubsection{It is Harder for ICL to Improve the Learning Gain on Stronger Model}
        \label{subsubsec:information_gain}
    
        Apart from the slope, the intercept of the fitted line also reflects the effectiveness of ICL under different settings.
        We statistic the intercept under different datasets and models, which is shown in Figure~\ref{fig:information_gain}.
        From the figure, we observe that as model capacity increases, the corresponding intercepts decrease, indicating that:
        \textit{(i)} From the perspective of the learning gain, the intercept reflects the overall magnitude of learning gain attributed to demonstrations for a given model and task, where a smaller intercept suggests less learning gain.
        \textit{(ii)} From the perspective of error estimation (Appendix~\ref{app:error_of_metric}), a smaller intercept implies a smaller discrepancy between \( p \) and \( \hat{p} \), meaning that the empirical predictor more closely approximates the oracle predictor.
        In summary, as model capacity increases, model predictions become more aligned with the oracle predictor, but the overall learning gain from demonstrations also diminishes.

\subsection{RQ3. Can Synthetic Data Accurately Reflect the ICL Effectiveness?}
    \begin{table}[t]
        \centering
        \small
        \caption{
            Experiment results of Llama2-7b and Llama3.1-8b on MATH and Amazon Review using labeled data (w. Human) and synthesized data (w/o. Human).
            $\Delta$ denotes the performance change with ICL.
            $r_{\hat{p}}x + b$ represents the fitted line between the contextual relevance and the learning gain.
        }
        \begin{tabular}{ll|cl|cl}
    \toprule
    \multirow{2}{*}{\textbf{Dataset}} & \multirow{2}{*}{\textbf{Data}} & \multicolumn{2}{c|}{\textbf{Llama2-7b}} & \multicolumn{2}{c}{\textbf{Llama3.1-8b}} \\
     & & $\Delta$ & $r_{\hat{p}}x + b$ & $\Delta$ & $r_{\hat{p}}x + b$ \\
    \midrule
    \multirow{2}{*}{MATH} & w. Human & $+9.6$ & $1.03 x + 0.03$ & $+2.4$ & $0.34 x - 0.00$ \\
     & w/o. Human & $+6.0$ & $0.75x + 0.00$ & $+1.3$ & $0.16x+0.02$ \\
    \midrule
    \multirow{2}{*}{Amazon} & w. Human & $+0.5$ & $0.07 x - 0.06$ & $+5.0$ & $0.94 x - 0.09$ \\
     & w/o. Human & $+0.0$ & $0.03x + 0.02$ & $+4.0$ & $0.42x-0.20$ \\
    \bottomrule
\end{tabular}

        \label{tab:synthesis_result}
    \end{table}

    To verify the conclusions regarding the computation of \ourmetric for synthetic data presented in \S\ref{subsec:metric_without_label}, we conduct experiments to calculate \ourmetric with synthetic data. 
    The experimental results are shown in Table~\ref{tab:synthesis_result}.
    From the table, we observe the following:
    \textit{(i)} The trend of \ourmetric using synthetic data is consistent with that derived from labeled data, demonstrating that synthetic data can effectively reflect the ICL effectiveness;
    \textit{(ii)} Compared to labeled data, the values of \ourmetric obtained from synthetic data are smaller, which supports the conclusion of Theorem~\ref{the:metric_without_labeling}.
    Therefore, based on the discussion in \S\ref{subsec:main_experiment}, in data-insufficient scenarios, if the synthetic data yields $\ourmetric > 0.2$\, which is the empirical LCS threshold for effective ICL using labeled data, we can conclude that ICL is effective for the given model and dataset, guiding the practitioners to decide whether to label the data.

    \section{Related Work}
        \subsection{In-Context Learning}
    In-context learning guides the LLM reasoning process by providing several task-relevant demonstrations in the input, thereby improving performance \cite{brown-etal-2020-gpt3,dong-etal-2024-ICLsurvey,zhao2025is}. 
    Existing ICL research can be broadly categorized into two main areas: constructing high-quality demonstrations and improving demonstration selection performance. 
    For demonstration construction, many works focus on the offline enhancement of demonstration quality. 
    This includes methods aimed at increasing demonstration diversity, for instance, by generating synthetic data tailored to a given task or by selecting diverse demonstrations to improve compositional generalization \cite{wang-etal-2024-improving-demonstration,wang2025incontext,chen-etal-2023-self,su-etal-2024-demonstration,levy2022diverse}. 
    Another key aspect of offline construction is synthesizing or augmenting reasoning steps within existing demonstrations to better guide the inference process \cite{li-etal-2024-self-prompting,zelikman2022star,zhao2023selfexplain}. 
    Other methods focus on the online synthesis of demonstrations, where demonstrations are generated or rewritten dynamically based on the user input to enhance reasoning performance, sometimes even leveraging the LLM itself to create these demonstrations \cite{he-etal-2024-self,chang-fosler-lussier-2023-selective,kim2022selfgenerated}. 
    In the domain of demonstration selection, research primarily explores how to choose demonstrations most relevant to the user query, with some approaches also incorporating active learning principles to identify the most informative demonstration \cite{luo2024incontext,vu2023active}. 
    Selection strategies include those based on n-grams~\cite{li-etal-2023-unified}, semantic similarity using embeddings~\cite{yang-etal-2023-representative,luo2023dricl}, or hybrid methods that combine multiple diverse strategies for retrieval and ranking \cite{wan2025from,wang-etal-2024-learning,hao2022structuredpromptingscalingincontext}.

    However, despite the improvements achieved by the above approaches, recent studies have shown that ICL does not consistently yield performance improvement across various models and tasks, rendering ICL ineffective \cite{deepseekai2025deepseekr1}. 
    Therefore, in this paper, we investigate how to evaluate the ICL effectiveness for a given task and model, informing the design and application of future ICL methods.

\subsection{Mechanism Analysis of In-Context Learning}
    Many studies have investigated the mechanisms underlying ICL to leverage ICL capabilities, improving reasoning performance \cite{zhou-etal-2024-mystery-survey,dong-etal-2024-ICLsurvey}. 
    One line of research explores the mechanism of ICL by controlling the types of tasks used during pretraining \cite{edelman2024the,han-etal-2023-understanding}. 
    Current mainstream work suggests that ICL ability arises from task diversity rather than data scale, with models gradually generalizing from solving in-domain tasks to solving out-of-domain tasks \cite{raventos2023pretraining}.
    Additionally, some studies find that the modules responsible for knowledge acquisition and ICL ability are functionally independent \cite{nguyen2025differential}.
    Increasing the amount of data primarily enhances the knowledge-related components, while improvements in ICL depend more on the diversity of tasks encountered during training.
    Another line of work focuses on ICL reasoning, aiming to discover the ICL mechanism by examining the relationship between provided demonstrations and the user question \cite{park2025competition,li2025llmseasilylearnreason,min-etal-2022-rethinking,wang2023label}.
    Some studies argue that models perform ICL by learning the mapping between inputs and labels in the demonstrations, thereby improving task-solving performance \cite{kossen2024incontext}.
    Other research suggests that models learn the reasoning process embedded in the demonstrations and enhance reasoning performance by understanding and mimicking these processes \cite{lampinen-etal-2022-language}.

    However, the aforementioned studies mainly focus on explaining the mechanism of ICL, often presuming that ICL is inherently effective. 
    In contrast, recent studies have shown that ICL does not lead to performance improvement on certain tasks and models \cite{deepseekai2025deepseekr1,huang2025explainable,zheng2025cursecotlimitationschainofthought}.
    Therefore, in this work, we investigate the main factors influence the ICL effectiveness and propose the metric to evaluate the ICL effectiveness, to inform and inspire future research.

    \section{Conclusion}
        In this paper, we propose a novel metric \ourmetric, to evaluate the ICL effectiveness. 
        \ourmetric overcomes the low reliability and poor attribution issues of performance-based metrics by measuring the variation in the learning gain with the contextual relevance. 
        Based on \ourmetric, we first discuss two primary factors that contribute to poor ICL effectiveness: poor contextual alignment capability and strong output calibration capability, demonstrating the strong attribution of \ourmetric. 
        Analytical experiments show that \ourmetric can effectively reflect the effectiveness of ICL even on demonstrations where ICL does not lead to performance improvements, indicating high reliability. 
        Furthermore, we present that the results of \ourmetric on synthetic data are lower than those on real data, to inspire the application of \ourmetric in data-insufficient scenarios.

    \newpage
    \bibliography{references}
    \bibliographystyle{plain}
    
    \medskip
    \newpage
    \appendix
    \section{Limitations and Ethics}
    \label{app:limitation_and_ethic}

    \subsection{Limitations}
        \textit{(i)} The current experimental datasets and models are limited, where future work will validate \ourmetric on a broader range of models and datasets.
        \textit{(ii)} Although we discuss that contextual alignment and output calibration capabilities are key factors influencing the ICL effectiveness, the underlying factors that affect these two capabilities warrant further investigation.

    \subsection{Ethics}
        All datasets and models used in this paper are publicly available, and our usage follows their licenses and terms.

\section{Prove}
    \subsection{Equation~\ref{equ:loss_function_decomposed}}
        \label{app:prove_of_loss_function}

        \begin{proof}
            Suppose $X = (x_1, ..., x_{|X|})$, where $x_i$ is the token of $X$, we can derive that:
            \begin{align*}
                \mathbb{L}_{p}(X|K;D;Q) &= -\log p(X|K;D;Q) \\
                &= \sum_{t=0}^{T} \left( -\log p(x_t|D;Q;x_{1:t-1}) \right) \\
                &= \sum_{t=0}^{T} \left( -\log \left( \frac{p(x_t|Q;x_{1:t-1}) p(D|Q;x_{1:t})}{p(D|Q;x_{1:t-1})} \right) \right) \\
                &= \mathbb{L}_{p}(X|K;Q) - \sum_{t=0}^{T} \left( \log \left( \frac{p(D|Q;x_{1:t})}{p(D|Q;x_{1:t-1})} \right) \right) \\
                &= \mathbb{L}_{p}(X|Q) - \left( \log p(D|Q;X) - \log p(D|Q) \right)
            \end{align*}
        \end{proof} 

    \subsection{Theorem~\ref{the:metric}}
        \label{app:prove_of_metric}

        \begin{proof}
            \begin{align*}
                p(X|Q;D) - p(X|Q) 
                &= \frac{p(X,Q,D)}{p(Q,D)} - \frac{p(X,Q)}{p(Q)} \\
                &= \frac{p(X,Q,D)p(Q) - p(X,Q)p(Q,D)}{p(Q,D)p(Q)} \\
                &= \frac{p(D|Q,X)p(Q,X)p(Q) - p(X,Q)p(D|Q)p(Q)}{p(Q,D)p(Q)} \\
                &= \frac{p(X|Q)}{p(D|Q)} \left( p(D|Q,X) - p(D|Q) \right) \\
                &= \frac{p(X|Q)}{p(D|Q)} I(X \rightarrow D|Q)
            \end{align*}
                
            Therefore, we can conclude that:
            \[
            I(X \rightarrow D|Q) = \frac{p(D|Q)}{p(X|Q)} I(D \rightarrow X|Q)
            \]
        \end{proof}

    \subsection{Error of Theorem~\ref{the:metric}}
        \label{app:error_of_metric}

        Assuming the error of the empirical predictor relative to the true predictor is \(\hat{p}(A|B) = p(A|B) + \varepsilon (A|B)\), where $A, B$ are any random variables.
        We suppose that \(r_p \geq \frac{\varepsilon(D|Q)}{\varepsilon(X|Q)} \geq \frac{\varepsilon(D|Q;X)}{\varepsilon(X|Q;D)}\), i.e., the error growth rate with introduced demonstrations is smaller than that without demonstrations, which is further smaller than the ICL effectiveness.
        According to Theorem~\ref{the:metric}, the slope of the fitted line can be approximated as:
        
        \[
        \frac{I_{\hat{p}}(D|Q;X)}{I_{\hat{p}}(X|Q;D)} = \frac{(p(D|Q;X) - p(D|Q)) + (\varepsilon(D|Q;X) - \varepsilon(D|Q))}{(p(X|Q;D) - p(X|Q)) + (\varepsilon(X|Q;D) - \varepsilon(X|Q))}
        \]
        
        Direct computation yields:
        
        \[
        \frac{\hat{p}(D|Q)}{\hat{p}(X|Q)} = \frac{p(D|Q) + \varepsilon(D|Q)}{p(X|Q) + \varepsilon(X|Q)}
        \]
        
        Thus, we have:
        
        \[
        \Delta_I := \frac{I_{\hat{p}}(D|Q;X)}{I_{\hat{p}}(X|Q;D)} - \frac{I_p(D|Q;X)}{I_p(X|Q;D)} = \frac{\varepsilon(D|Q;X) - \varepsilon(X|Q;D)r_p}{I_p(X|D;Q)\left(I_p(X|D;Q) + \varepsilon(X|Q;D)\right)}
        \]
        
        \[
        \Delta_p := \frac{\hat{p}(D|Q)}{\hat{p}(X|Q)} - \frac{p(D|Q)}{p(X|Q)} = \frac{\varepsilon(D|Q;X) - \varepsilon(X|Q;D)r_p}{p(X|D;Q)\left(p(X|D;Q) + \varepsilon(X|Q;D)\right)}
        \]
        
        Assuming \(I_p(X|D;Q) \leq p(X|D;Q)\), i.e., the information the model learns about \(D\) from \(X\) is less than the information inherently contained in the model, we have:
        
        \[
        \Delta_I \leq \Delta_p
        \]
        
        This implies that using the slope as a metric for ICL effectiveness has a smaller error compared to using \(\frac{\hat{p}(D|Q)}{\hat{p}(X|Q)}\).

    \subsection{Theorem~\ref{the:metric_without_labeling}}
        \label{app:prove_of_metric_without_labeling}

        \begin{proof}
            Since $\hat{X}=\argmax_{X \sim \mathcal{X}} \hat{p}(X|Q)$, we can conclude that $\hat{p}(\hat{X}|Q) \geq \hat{p}(X^*|Q)$.
            Based on the total probability theorem, we can draw that:
            $$\hat{p}(\hat{D}|Q) = \sum_{X \sim \mathcal{X}} \hat{p}(\hat{D}|Q;X)$$
            $$\hat{p}(D^*|Q) = \sum_{X \sim \mathcal{X}} \hat{p}(D^*|Q;X)$$
            Considering that $\hat{p}(X | Q; \hat{D}) \leq \hat{p}(X | Q; D^*), \forall X \sim \mathcal{X},Q \sim \mathcal{Q}$, it can be concluded that $\hat{p}(\hat{D}|Q) \leq \hat{p}(D^*|Q)$.
            Therefore, we can draw the conclusion that:
            $$\frac{\hat{p}(\hat{D}|Q)}{\hat{p}(\hat{X}|Q)} \leq \frac{\hat{p}(D^*|Q)}{\hat{p}(X^*|Q)}$$
        \end{proof}

\section{Calculation of \ourmetric}
    \label{app:metric_calculation}

    In this section, we present how to calculate \ourmetric, which primarily involves two sequential steps: reasoning process paraphrasing and likelihood calculation.
    The prompts employed for these computations are detailed in Appendix~\ref{app:prompt}.

    The reasoning process paraphrasing step requires models to restructure human-labeled reasoning process according to their preferred reasoning style when provided with given reasoning process. 
    This adaptation is crucial because discrepancies between human-labeled reasoning formats and model-preferred reasoning patterns (e.g., <think> tag of Llama-R1 \cite{deepseekai2025deepseekr1}) could lead to inflated information gain measurements that reflect stylistic variations rather than knowledge acquisition.
    To mitigate this confounding factor, we implement reasoning process paraphrasing to eliminate format-induced biases, thereby ensuring that computational results authentically reflect knowledge-derived information learned from demonstrations. 
    Specifically, for each data instance and demonstration, we input the question, answer, and human-labeled reasoning process (if provided), instructing the model to rephrase the output using its preferred reasoning style.

    Following the paraphrasing, we calculate the likelihood with paraphrased results. 
    For conditional probabilities expressed as $\hat{p}(A|B)$, we treat $B$ as user input and $A$ as model output, encapsulating them into a formatted string using the model chat template.
    This composite string is then processed through the model to obtain token-level likelihoods. 
    The joint likelihood of sequence $A$ is computed by multiplying the probabilities of all constituent tokens. 
    To minimize the confounding effects of sequence length on probability comparisons, we apply length normalization to all computed likelihood values \cite{dai2019transformerxlattentivelanguagemodels}.
    This standardized approach ensures fair comparison across outputs of varying lengths while preserving the probabilistic relationships between different reasoning processes.

\section{Prompts}
    \label{app:prompt}

    In the section, we introduce the used prompts of this paper.
    The reasoning prompts of \S\ref{sec:experiment} can be seen in \citep{chen2023program,grattafiori2024llama3,deepseekai2025deepseekr1}.
    The prompts used for the paraphrasing and the synthesis are shown in Table~\ref{tab:prompt_paraphrase} and Table~\ref{tab:prompt_synthesis}.

    \begin{table}[ht]
        \centering
        \small
        \caption{
            The prompt of the paraphrase.
        }
        \begin{tabular}{p{0.9\linewidth}}
    \toprule
    \textbf{Prompt of Paraphrasing} \\
    \midrule
    <Begin of Task Definition> \\
    \{definition\} \\
    <End of Task Definition> \\
    <Begin of Input> \\
    \{question\} \\
    <End of Input> \\
    <Begin of Hint> \\
    \{hint\} \\
    <End of Hint> \\
    <Begin of Answer> \\
    \{answer\} \\
    <End of Answer> \\
     \\
    Considering the above task definition, generate the reasoning process of the given input and answer with the hint (could be empty). \\
    \bottomrule
\end{tabular}
        \label{tab:prompt_paraphrase}
    \end{table}

    \begin{table}[ht]
        \centering
        \small
        \caption{
            The prompt of the demonstration synthesis.
        }
        \begin{tabular}{p{0.9\linewidth}}
    \toprule
    \textbf{Prompt of Synthesis} \\
    \midrule
    \textasciigrave \textasciigrave \textasciigrave md \\
    \{task\_definition\} \\
    \textasciigrave \textasciigrave \textasciigrave \\
    Given Question: \{question\} \\
    \\
    Based on the above task definition and the given question, synthesize a question and the corresponding answer that is similar to the given question of the task. \\
    \bottomrule
\end{tabular}

        \label{tab:prompt_synthesis}
    \end{table}

\section{The Factors that Affect ICL Effectiveness}
    \label{app:effective_factor}

    Following the discussion of \S\ref{subsec:metric}, in this section, we delve deeper into the factors that influence the ICL effectiveness, specifically the meaning of \( \hat{p}(D|Q) \) and \( \hat{p}(X|Q) \). 
    Our primary focus is on different predictors \( \hat{p}_1 \) and \( \hat{p}_2 \) applied to the same data, assuming that the answer \( X = \arg\max_{X \in \mathcal{X}} p(X|Q) \) is the correct answer, and the demonstration \( D = \arg\max_{D \in \mathcal{D}} p(D|Q) \) is the most relevant demonstration to the question \( Q \).

    \paragraph{Contextual Alignment Capability \bm{$\hat{p}(D|Q)$}}
        If \( \hat{p}_1(D|Q) \geq \hat{p}_2(D|Q) \), it indicates that \( \hat{p}_1 \) has a stronger ability to judge the relevance of demonstrations to the question compared to \( \hat{p}_2 \), showing that \( \hat{p}_1 \) is a better demonstration selector.
        From the perspective of demonstrations, this means that \( \hat{p}_1 \) is better at understanding the information in the demonstration and determining its relationship with the user question \( Q \), reflecting that \( \hat{p}_1 \) has a stronger ICL ability than \( \hat{p}_2 \).

        It is worth noting that while both $\hat{p}(D|Q)$ and $I(D \rightarrow X | Q)$ measure the consistency between the demonstration and the user input to some extent, their fundamental perspectives differ. 
        $I(D \rightarrow X | Q)$ primarily focuses on the data perspective, measuring the relevance between the input and the demonstration under the assumption that the model is an oracle. 
        In contrast, $\hat{p}(D|Q)$ primarily focuses on the model perspective, observing whether the model has the capability to gauge the relevance between the input and the demonstration, assuming that the demonstration is highly relevent to the user input.

    \paragraph{Output Calibration Capability \bm{$\hat{p}(X|Q)$}}
        If \( \hat{p}_1(X|Q) \geq \hat{p}_2(X|Q) \), it implies that \( \hat{p}_1 \) has a stronger ability to judge the correct answer compared to \( \hat{p}_2 \), meaning that \( \hat{p}_1 \) is a better answer scorer.
        It should be noticed that \( \hat{p}(X|Q) \) does not directly reflect the model ability to solve the given question. 
        This is because the model generates answers using greedy decoding, which means the generated answer could not be the answer with the highest likelihood. 
        Rather, \( \hat{p}(X|Q) \) represents the score the model assigns to a given answer, reflecting the model ability to assess the consistency between the answer and the question.

\section{Detail of Benchmarks}
    \label{app:benchmark}

    \begin{table}[ht]
        \centering
        \small
        \caption{
            The scales of test set and demonstrations of each dataset.
        }
        \begin{tabular}{l|cc}
            \toprule
            \textbf{Dataset} & \textbf{Test Set} & \textbf{Demonstration} \\
            \midrule
            GSM8K & $1319$ & $7473$ \\
            MATH & $500$ & $7496$ \\
            HumanEval & $164$ & $596$ \\
            MBPP & $378$ & $596$ \\
            ARC-Challenge & $1172$ & $1119$ \\
            MMLU-Pro & $1000$ & $70$ \\
            FinQA & $1147$ & $6251$ \\
            Amazon Review & $200$ & $1800$ \\
            \bottomrule
        \end{tabular}
        \label{tab:dataset_scale}
    \end{table}

    In this section, we discuss the datasets we used in this paper in detail.
    The scale of the test set and the demonstrations of each dataset are shown in Table~\ref{tab:dataset_scale}.

    \paragraph{GSM8K}
        GSM8K~\cite{cobbe2021gsm8k} is a high-quality dataset consisting of grade-school level math problems. 
        We directly use the training set as the demonstration pool.

    \paragraph{MATH}
        MATH~\cite{hendrycks2021math} is a dataset of high school competition-level math problems covering various domains, such as algebra, probability, and geometry. 
        Following \citep{lightman2024lets}, we use a sampled subset of $500$ examples for evaluation. 
        We use the training set as the demonstration pool.

    \paragraph{HumanEval}
        HumanEval~\cite{chen2021humaneval} is a Python-based code generation benchmark. 
        We follow the evaluation protocol of \citep{evalplus}. 
        Since the dataset does not provide a labeled training set, we use demonstrations from MBPP as the demonstration pool.
    
    \paragraph{MBPP}
        MBPP~\cite{austin2021mbpp} is another Python-based code generation benchmark. 
        Compared to HumanEval, it is larger in scale and includes a split between validation and test sets. 
        In this paper, we adapt the evaluation on the test set and use the remaining data as the demonstration pool, following the evaluation protocol of \citep{evalplus}.
    
    \paragraph{ARC-Challenge}
        ARC-Challenge~\cite{yadav-etal-2019-arc} is a difficult question-answering dataset focusing on scientific knowledge. 
        We directly use the training set as the demonstration pool.
    
    \paragraph{MMLU-Pro}
        MMLU-Pro~\cite{wang2024mmlupro} is a multi-task benchmark designed to comprehensively evaluate LLMs on professional domain knowledge and complex reasoning capabilities. 
        As the dataset only provides validation and test sets, we use the validation set as the demonstration pool and evaluate on the test set.

    \paragraph{FinQA}
        FinQA~\cite{chen2021finqa} is a question-answering dataset in the financial domain.
        It requires models to perform numerical reasoning and calculations based on given financial tables and textual information. 
        We use the training set as the demonstration pool.
    
    \paragraph{Amazon Review}
        The Amazon~Review~\cite{ni-etal-2019-amazon} dataset consists of a large number of user ratings and textual reviews on products from the Amazon platform, and it is widely used in sentiment analysis and recommendation system research. 
        Due to the large scale of the dataset, we select the \textit{Health and Personal Care} category as the test set and use \textit{All Beauty}, \textit{Digital Music}, and \textit{Software} as the demonstration pool.

\section{Main Experiment Results}
    \label{app:experiment_result}

    \subsection{Overall Perfromance}
        \begin{table}[ht]
            \centering
            \small
            \caption{
                The performance of different models on different datasets using 0-shot and 1-shot.
                $\Delta$ is the performance change of 1-shot compared with 0-shot.
                HumanE denotes HumanEval, A-C denotes ARC-Challenge, M-P denotes MMLU-Pro, and Amazon denotes Amazon Review.
            }
            \begin{tabular}{llcccccccc}
    \toprule
    \multirow{2}{*}{\textbf{Model}} & \multirow{2}{*}{\textbf{Shot}} & \multicolumn{2}{c}{\textbf{Math}} & \multicolumn{2}{c}{\textbf{Code}} & \multicolumn{2}{c}{\textbf{Reason}} & \multicolumn{2}{c}{\textbf{Domain}} \\
    \cmidrule(lr){3-4} \cmidrule(lr){5-6} \cmidrule(lr){7-8} \cmidrule(lr){9-10}
    & & GSM8K & MATH & HumanE & MBPP & A-C & M-P & FinQA & Amazon \\
    \midrule
    \multirow{3}{*}{Llama2-7b} & 0 & $12.7$ & $5.0$ & $14.0$ & $23.0$ & $34.6$ & $14.1$ & $10.5$ & $28.5$ \\
    & 1 & $27.7$ & $14.6$ & $13.4$ & $22.5$ & $46.2$ & $19.6$ & $17.8$ & $29.0$ \\
    \cmidrule{2-10}
    & $\Delta$ & $+10.0$ & $+9.6$ & $-0.6$ & $-0.5$ & $+11.6$ & $+5.5$ & $+7.3$ & $+0.5$ \\
    \midrule
    \multirow{3}{*}{Llama3.1-8b} & 0 & $86.4$ & $48.4$ & $65.9$ & $54.8$ & $82.1$ & $50.4$ & $49.7$ & $63.5$ \\
    & 1 & $84.2$ & $50.8$ & $63.4$ & $55.6$ & $80.2$ & $53.0$ & $54.6$ & $68.5$ \\
    \cmidrule{2-10}
    & $\Delta$ & $-1.8$ & $+2.4$ & $-2.5$ & $+0.8$ & $-1.9$ & $+2.6$ & $+4.9$ & $+5.0$ \\
    \midrule
    \multirow{3}{*}{Llama-R1-8b} & 0 & $86.1$ & $75.4$ & $70.7$ & $67.2$ & $84.8$ & $58.2$ & $45.2$ & $53.5$ \\
    & 1 & $80.1$ & $74.2$ & $67.7$ & $60.8$ & $84.5$ & $52.7$ & $43.4$ & $65.0$ \\
    \cmidrule{2-10}
    & $\Delta$ & $-6.0$ & $-1.2$ & $-3.0$ & $-6.4$ & $-0.3$ & $-5.5$ & $-1.8$ & $+11.5$ \\
    \bottomrule
\end{tabular}

            \label{tab:performance}
        \end{table}
    
        In this section, we present the performance of 0-shot and 1-shot on different models and datasets, as shown in Table~\ref{tab:performance}.

    \subsection{Figurative Illustrations of Learn-to-Context Slope}        
\begin{figure}
            \centering
            \small
            \input{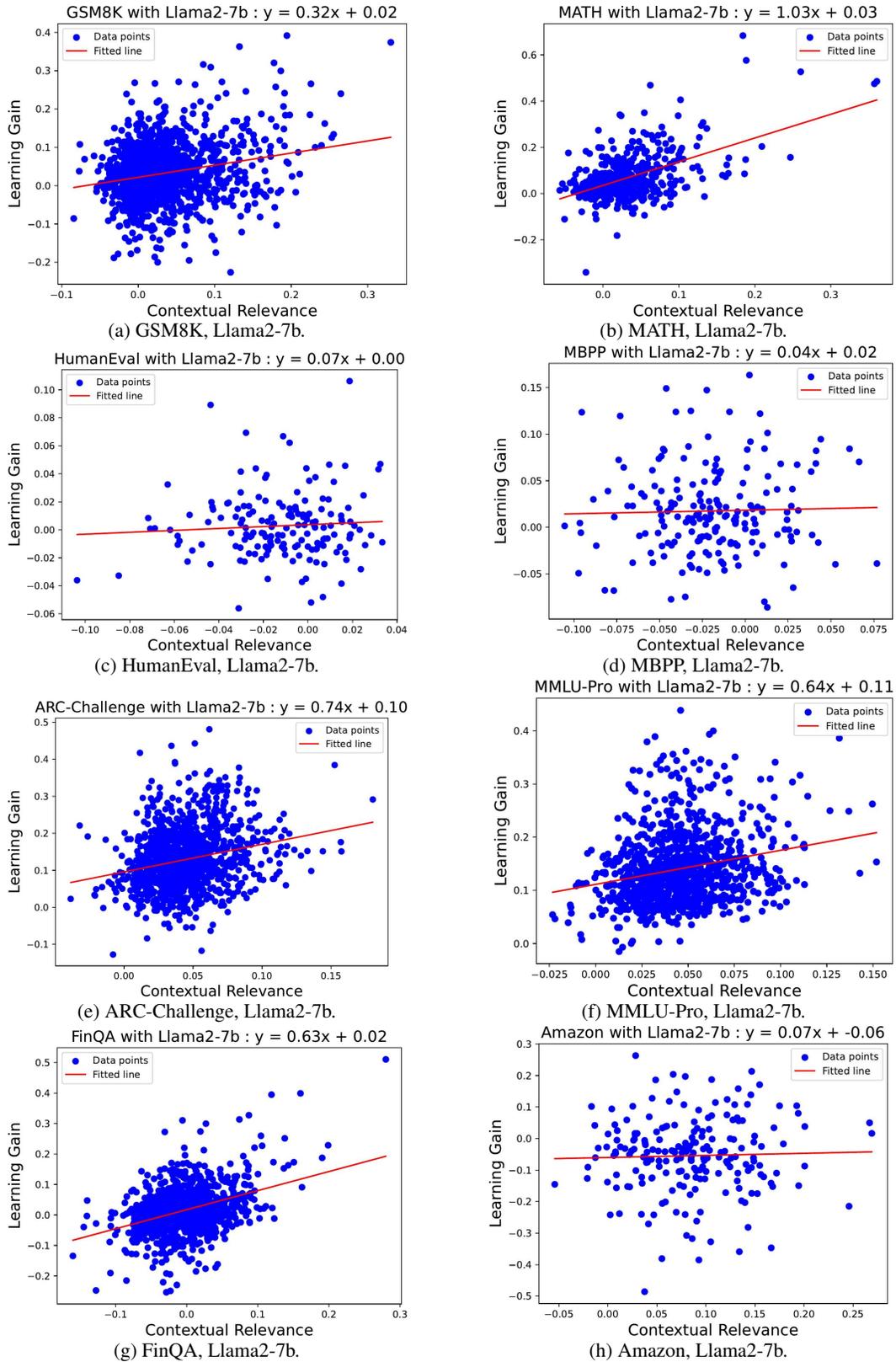}
            \caption{
                The variation of $I_{\hat{p}}(X \rightarrow D | Q)$ (y-axis) with $I_{\hat{p}}(D \rightarrow X | Q)$ (x-axis) on different datasets using Llama2-7b.
                The title of each plot displays the corresponding dataset and fitted line.
                Each \textblue{blue} dot in the plot represents a data point, and the \textred{red} line indicates the fitted line of the data points.
            }
            \label{fig:llama2_7b_figure}
        \end{figure}

\begin{figure}
            \centering
            \small
            \input{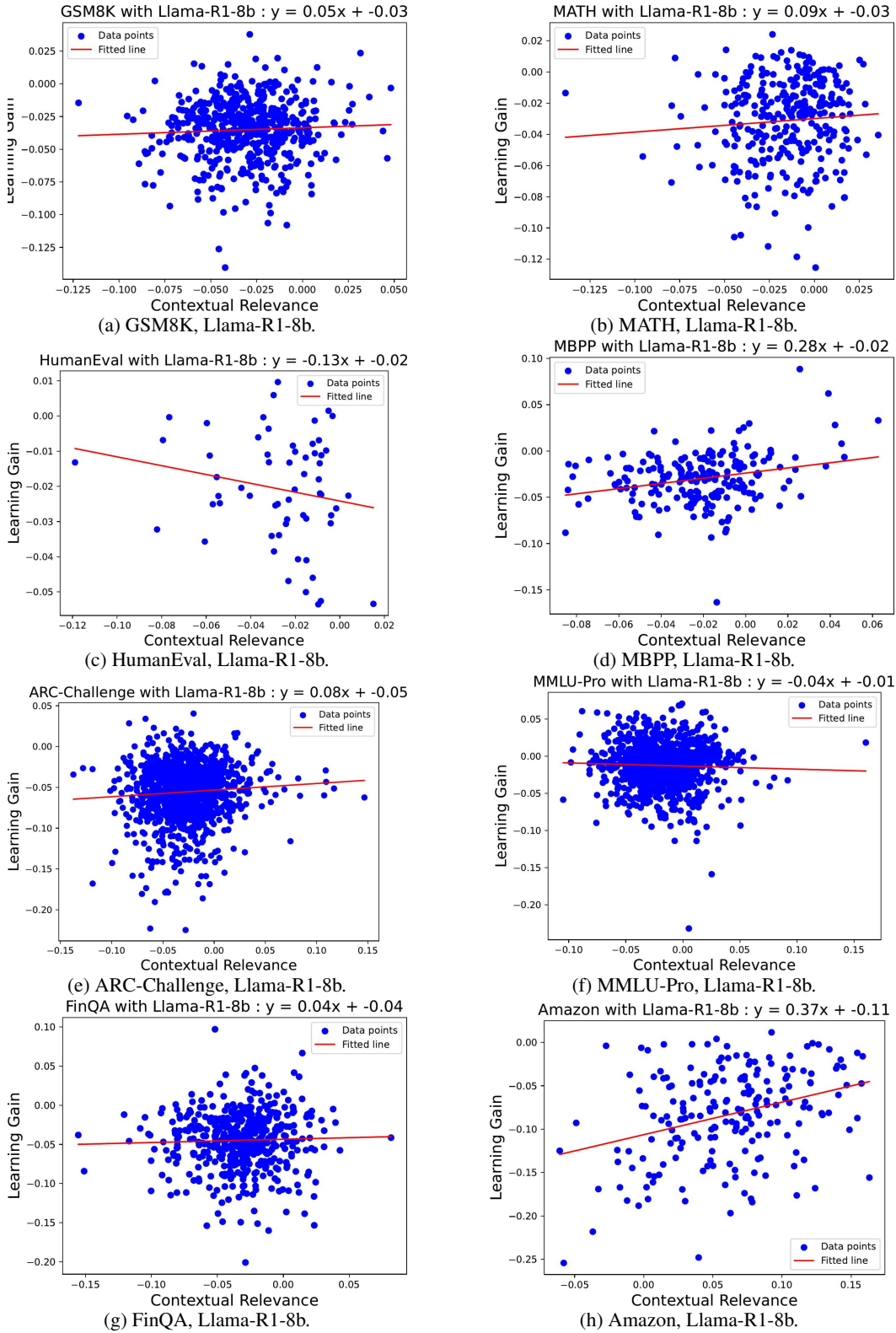}
            \caption{
                The variation of $I_{\hat{p}}(X \rightarrow D | Q)$ (y-axis) with $I_{\hat{p}}(D \rightarrow X | Q)$ (x-axis) on different datasets using Llama-R1-8b.
                The legend is the same as Figure~\ref{fig:llama2_7b_figure}.
            }
            \label{fig:llama_r1_8b_figure}
        \end{figure}

        In this section, we present the variation of $I_{\hat{p}}(X \rightarrow D | Q)$ with respect to $I_{\hat{p}}(D \rightarrow X | Q)$ under different settings, as illustrated in Figure~\ref{fig:llama2_7b_figure}, Figure~\ref{fig:llama3.1_8b_figure}, and Figure~\ref{fig:llama_r1_8b_figure}. 
        Considering that the number of data points could vary slightly across different models due to the potential for excessively long responses from certain models (e.g., Llama-R1-8b could persist in "thinking").

        \begin{figure}
            \centering
            \small
            \input{fig/curves/figure_llama3.1_8b}
            \caption{
                The variation of $I_{\hat{p}}(X \rightarrow D | Q)$ (y-axis) with $I_{\hat{p}}(D \rightarrow X | Q)$ (x-axis) on different datasets using Llama3.1-8b.
                The legend is the same as Figure~\ref{fig:llama2_7b_figure}.
            }
            \label{fig:llama3.1_8b_figure}
        \end{figure}

\section{Additional Experiments}
    \subsection{Different Similarity Measurement}
        \label{app:similarity_measurement}

        \begin{figure}
            \centering
            \small
            \input{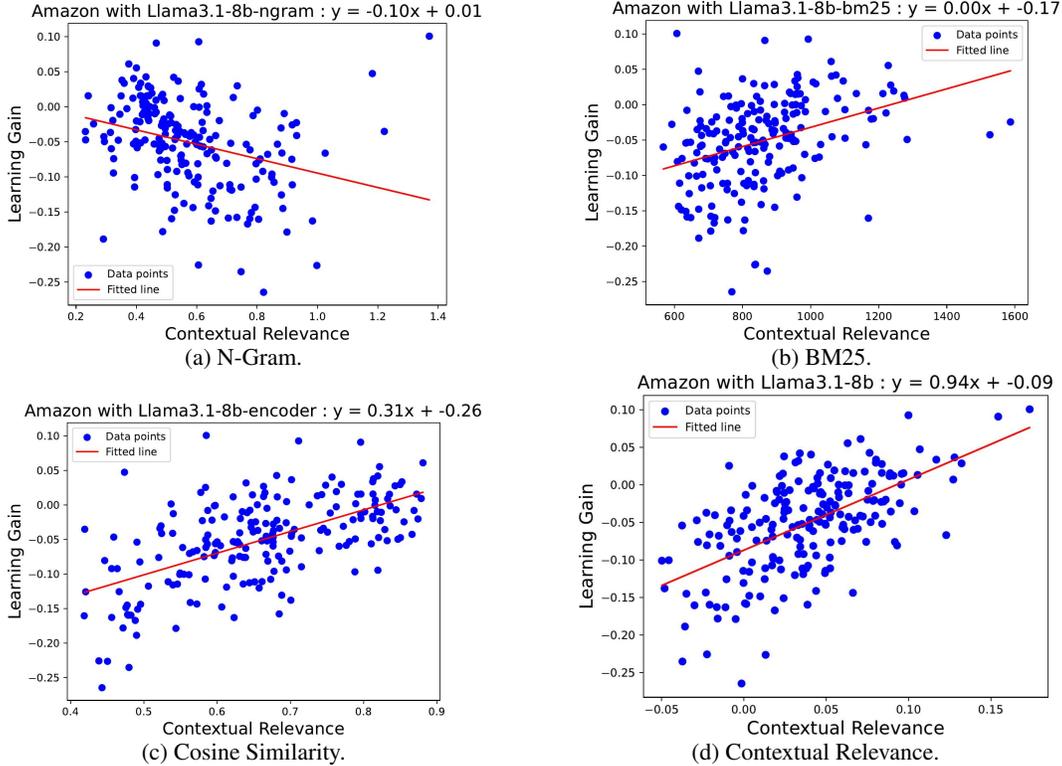}
            \caption{
                The variation of the learning gain (y-axis) with different similarity metrics (x-axis) on Amazon using Llama3.1-8b.
            }
            \label{fig:similarity}
        \end{figure}

        In this section, we discuss the impact of replacing the contextual relevance $I_{\hat{p}}(D \rightarrow X | Q)$ with other metrics. 
        We conduct experiments on Llama3.1-8b using the Amazon Review dataset, where we replace the similarity measure with n-gram~\cite{ngram}, BM25~\cite{robertson-etal-2009-bm25}, and cosine similarity~\cite{cosine} to evaluate the similarity between the provided demonstration and user input. 
        The experimental results are shown in Figure~\ref{fig:similarity}. 
        From the figure, we observe the following:
        \textit{(i)} For effective similarity measures (e.g., BM25, cosine similarity), the observed ICL effectiveness is consistent with using the contextual relevance;
        \textit{(ii)} However, for metrics with poorer performance (e.g., n-gram), the ICL effectiveness is not accurately reflected, demonstrating that n-gram fails to properly capture the similarity between demonstrations and user inputs.

    \subsection{Different Model}
        \label{app:experiment_model_type}

        \begin{table}[ht]
            \centering
            \small
            \caption{
                Performance and fitted lines across different models and datasets.
                ARC-C denotes ARC-Challenge, and Amazon denotes Amazon Review.
                $\Delta$ denotes the performance change of 1-shot relative to 0-shot, where performance gains $< 1.0$ are marked in \textred{red}.  
                $r_{\hat{p}}x + b$ represents the fitted line with $I_{\hat{p}}(X \rightarrow D | Q)$ as the x-axis and $I_{\hat{p}}(D \rightarrow X | Q)$ as the y-axis, where $r_{\hat{p}}$ values $< 0.2$ are highlighted in \textred{red}. 
            }
            \begin{tabular}{l|cc|cc|cc}
    \toprule
    \multirow{2}{*}{\textbf{Model}} & \multicolumn{2}{c|}{\textbf{MATH}} & \multicolumn{2}{c|}{\textbf{FinQA}} & \multicolumn{2}{c}{\textbf{Amazon}} \\
     & $\Delta$ & $r_{\hat{p}}x + b$ & $\Delta$ & $r_{\hat{p}}x + b$ & $\Delta$ & $r_{\hat{p}}x + b$ \\
    \midrule
    Llama3.1-8b & $2.4$ & $0.34 x - 0.00$ & $4.9$ & $0.82 x - 0.06$ & $5.0$ & $0.94 x - 0.09$ \\
    Qwen2.5-7b & $2.2$ & $0.81x-0.16$ & $4.9$ & $0.29x - 0.09$ & $27.0$ & $0.42x+0.00$ \\
    Llama3.1-70b & \textred{$-3.6$} & \textred{$-0.13 x - 0.04$} & $7.6$ & $0.77 x - 0.07$ & $16.0$ & $0.79 x - 0.18$ \\
    \bottomrule
\end{tabular}

            \label{tab:different_model}
        \end{table}

        To evaluate the effectiveness of \ourmetric on the models with different scales and series, we adapt the experiments on Qwen2.5-7b~\cite{qwen2025qwen25technicalreport} and Llama3.1-70b~\cite{grattafiori2024llama3}.
        The experiment results are shown in Table~\ref{tab:different_model}.
        It can be seen that \ourmetric still reflects the ICL effectiveness on the models with different scales and series, proving the generalization of our metric.

    \subsection{Intercept under Different Model Scales}
        \label{app:intercept_scales}

        \begin{figure}
            \centering
            \small
            \begin{tikzpicture}[
  font=\small,
  x=0.02\textwidth,
  y=0.02\textwidth
]
  \useasboundingbox (-12, -12) rectangle (12, 12);

  \coordinate (origin) at (0,0);

  \def\n{8}                 
  \def\step{360/\n}         
  \def\labels{{"GSM8K","MATH","HumanEval","MBPP",
               "ARC-C","MMLU-Pro","FinQA","Amazon"}} 

  \def\ModelA{{10.00, 6.00, 6.67, 8.33, 3.33, 5.67, 4.0, 3.0}}
  \def\ModelB{{6.67, 4.67, 2.33, 5.0, 3.0, 5.33, 3.67, 2.0}}

  \def\maxRadius{10}      
  \def\labelRadius{12}    

  \foreach \r in {2,4,6,8,10} {
      \draw[gray!40] (origin) circle (\r);
  }

  \foreach \i in {1,...,\n} {
      \pgfmathtruncatemacro{\deg}{(\i-1)*\step}
      \draw (origin) -- (\deg:\maxRadius);
  }

  \foreach \i in {1,...,\n} {
      \pgfmathtruncatemacro{\deg}{(\i-1)*\step}
      \pgfmathparse{\labels[\i-1]}
      \let\dimName\pgfmathresult
      \node at (\deg:\labelRadius) {\dimName};
  }

  \foreach \i in {1,...,\n} {
      \pgfmathtruncatemacro{\j}{\i-1}

      \pgfmathparse{\ModelA[\j]}\let\valA\pgfmathresult
      \coordinate (A\i) at (\j*\step:\valA);

      \pgfmathparse{\ModelB[\j]}\let\valB\pgfmathresult
      \coordinate (B\i) at (\j*\step:\valB);
  }

  \draw[fill=blue!20, draw=blue, thick, opacity=0.5]
       (A1) \foreach \i in {2,...,\n}{--(A\i)} -- cycle;

  \draw[fill=red!20, draw=red, thick, opacity=0.5]
       (B1) \foreach \i in {2,...,\n}{--(B\i)} -- cycle;

  \node[blue, xshift=5, yshift=-10] at (A1) {Llama3.1-8b};
  \node[red, xshift=5,  yshift= 10] at (B1) {Llama3.1-70b};

\end{tikzpicture}
            \caption{
                The intercepts of the fitted lines of Llama3.1-8b and Llama3.1-70b.
            }
            \label{fig:information_gain_scale}
        \end{figure}
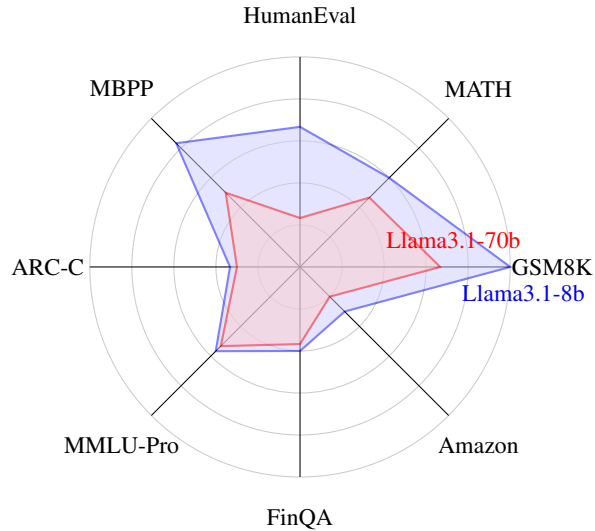

        To more thoroughly compare the differences in the effective information learned from demonstrations by LLMs of varying capabilities, we conduct experiments on LLMs of different scales within the same series. 
        The experimental results are shown in Figure~\ref{fig:information_gain_scale}. 
        From the figure, it can be observed that the intercept of Llama3.1-70b is generally smaller than that of Llama3.1-8b, as discussed in Section \S\ref{subsubsec:information_gain}, indicating that Llama3.1-70b learns less effective information.


\end{document}